\documentclass[10pt,twocolumn,letterpaper]{article}

\PassOptionsToPackage{table}{xcolor}

\usepackage[pagenumbers]{wacv}

\usepackage[accsupp]{axessibility}
\usepackage{multirow, graphicx, booktabs, colortbl}
\usepackage{amssymb}
\usepackage[table]{xcolor}
\definecolor{cbest}{RGB}{255,158,158}
\definecolor{csecond}{RGB}{255,205,155}
\definecolor{cthird}{RGB}{255,242,160}

\usepackage{longtable}
\usepackage{arydshln}
\usepackage{cuted}

\definecolor{cgt}{RGB}{255,176,0}

\newlength{\dsw}
\newlength{\dscolw}
\newlength{\dsrule}
\newlength{\dsrulegt}
\newlength{\dslab}
\newlength{\dsgap}
\newsavebox{\dsbox}

\newcommand{\dscell}[1]{%
  \fcolorbox{white}{white}{#1}}

\newcommand{\gtbadge}{%
  \colorbox{cgt}{\textcolor{black}{\small\bfseries\,GT\,}}}

\newcommand{\gtcell}[1]{%
  \sbox{\dsbox}{#1}%
  {\setlength{\fboxrule}{\dsrulegt}%
   \fcolorbox{cgt}{cgt}{%
    \rlap{\raisebox{\dimexpr\ht\dsbox-\totalheight-3pt\relax}[0pt][0pt]{%
          \hspace{3pt}\gtbadge}}%
    \usebox{\dsbox}}}}

\newcommand{\hd}[1]{\makebox[\dscolw][c]{\footnotesize #1}}
\newcommand{\hdn}[1]{\makebox[\dscolw][c]{\scriptsize #1}}
\newcommand{\rowlab}[1]{%
  \makebox[\dslab][c]{\rotatebox{90}{\scriptsize#1}}}

\newlength{\qlw}

\newcommand{\qlhd}[1]{\makebox[\qlw][c]{\scriptsize #1}}
\newcommand{\qlhdb}[2]{\makebox[\qlw][c]{\scriptsize\shortstack{#1\\#2}}}

\newlength{\dmw}
\newlength{\dmg}
\newlength{\dmi}
\newlength{\dmv}
\newlength{\dmb}
\newlength{\dmbh}
\newsavebox{\dmbox}
\newcommand{\dmhd}[1]{\makebox[\dmw][c]{\scriptsize #1}}
\newcommand{\dmbar}{%
  \sbox{\dmbox}{\includegraphics[width=5pt,height=\dmbh]{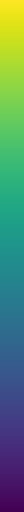}}%
  \makebox[\dmb][l]{\raisebox{0pt}[0pt][0pt]{%
    \usebox{\dmbox}%
    \rlap{\hspace{1.5pt}\raisebox{\dimexpr\ht\dmbox-1.15ex\relax}[0pt][0pt]{%
          \scriptsize$1$}}%
    \rlap{\hspace{1.5pt}\raisebox{0.15ex}[0pt][0pt]{\scriptsize$0$}}}}}

\definecolor{wacvblue}{rgb}{0.21,0.49,0.74}
\usepackage[pagebackref,breaklinks,colorlinks,allcolors=wacvblue]{hyperref}

\title{CHROMA-MILL: Joint Low-Light Enhancement and White Balance under Colored Illumination}

\author{
Pol Riubrogent-Comas\textsuperscript{1,2} \quad David Serrano-Lozano\textsuperscript{1,2} \quad Maria Pilligua\textsuperscript{1,2}\\
Michael S. Brown\textsuperscript{3} \quad Ramon Baldrich\textsuperscript{1,2} \quad Javier Vazquez-Corral\textsuperscript{1,2}\\
{ \small \textsuperscript{1}Computer Vision Center \quad \textsuperscript{2}Universitat Aut\`onoma de Barcelona \quad \textsuperscript{3}York University}\\[2pt]
{\small \url{https://color.cvc.uab.cat/chromamill/}}
}

\begin{document}
\maketitle
\setlength{\stripsep}{6pt}
\begin{strip}
    \vspace*{-16pt}
    \centering
    \includegraphics[width=0.95\linewidth]{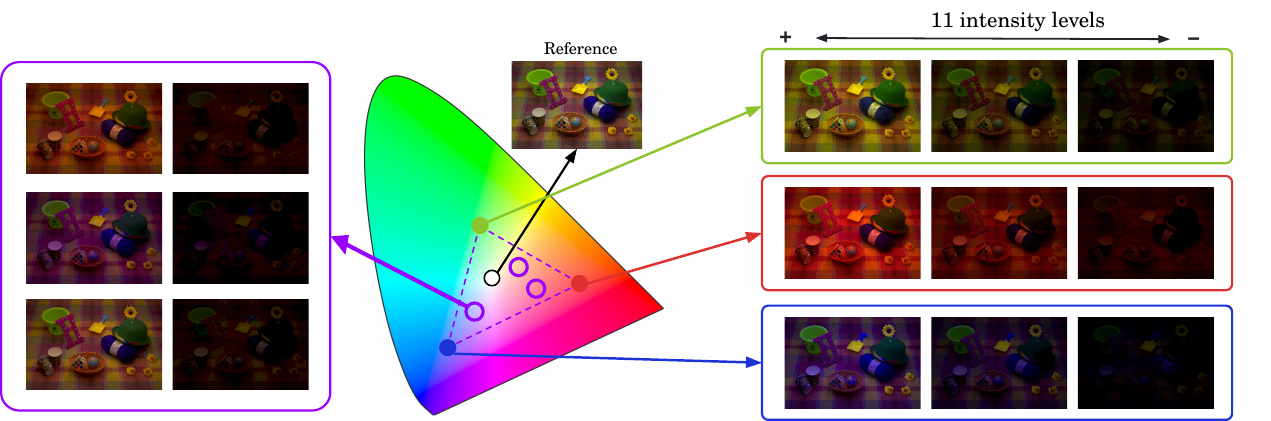}
    \par\vspace{8pt}
    \captionof{figure}{Overview of CHROMA-MILL. Scenes are captured under jointly varying illumination and intensity. The CIE~1931 diagram shows the four captured illuminants (three colored and one white); linear-RAW blending places the effective illuminant anywhere in their convex hull (dashed triangle). Each mixture is captured over 11 intensity levels, from well-lit to near-dark. The well-lit white-illuminant capture (Level~11, Reference) is the ground truth.}
    \label{fig:teaser}
\end{strip}

\begin{abstract}
Capturing an image in a low-light environment produces a weak, noisy sensor signal. Low-light image enhancement (LLIE) aims to restore it. Auto white balance (AWB) is also required for every image, and the same noisy signal degrades its ability to infer the scene illuminant. LLIE is typically performed first, followed by AWB, but this strategy remains largely unexamined because existing low-light datasets span a narrow range of illuminants and lack illuminant ground truth for evaluating AWB. We address this with CHROMA-MILL, a RAW dataset of 50 scenes captured with a DSLR and smartphone under four calibrated illuminants at eleven intensity levels. Because light is additive, we blend these captures in the linear RAW domain to synthesize physically valid illuminant mixtures with exact chromaticity ground truth. Benchmarking on this data, we find joint correction is best, and introduce CTI-Retinexformer, which supervises an illumination map, a chromaticity prediction, and an illumination-invariant scene embedding, outperforming the best cascade by $0.7$~dB PSNR$_L$ and $1.1~\Delta E_{00}$.
\end{abstract}

\section{Introduction}
\label{sec:intro}

Capturing an image in a low-light environment produces a weak sensor signal.
Reduced scene radiance lowers the signal-to-noise ratio, leaving captures that
are dark, noisy, and stripped of detail~\cite{chen2018sid,
pilligua2025evaluating}. Low-light image enhancement (LLIE) aims to restore
this signal, and deep models are now effective at brightening and
denoising~\cite{cai2023retinexformer, wei2018retinexnet, guo2020zerodce,
feijoo2025darkir, yan2025cidnet}.

Restoring the signal is not the only correction such an image needs. Auto
white balance (AWB) is required for every image a camera captures, and the same
noisy signal degrades it. An AWB algorithm infers the scene illuminant from
cues in the image itself, such as bright or near-neutral surfaces, and every
estimator---statistical or learned---needs those cues to be well
exposed~\cite{xie2026boosting}. In low light they are buried in the same noise
that corrupts the scene content, so the darker the capture, the less reliable
the illuminant estimate. The problem compounds because low-light capture
happens indoors and at night, where artificial sources impose color casts that
are both strong and varied: the illuminants hardest to estimate are the ones
most likely present when the signal is weakest.

LLIE is typically performed first, followed by AWB, but this strategy remains
largely unexamined. LLIE methods~\cite{cai2023retinexformer, guo2020zerodce,
feijoo2025darkir, yan2025cidnet} are scored on fidelity to a ground truth
carrying the input's own illuminant, so a residual color cast costs them
nothing; color constancy methods~\cite{buchsbaum1980, vandeweijer2007grayedge,
hu2017fc4, barron2017ffcc, afifi2020dwb} are scored on well-exposed inputs, and
their behavior under severe underexposure is almost
unexplored~\cite{lee2026rlawb}. Neither evaluation measures the pipeline the
two are deployed in.

The main issue is the lack of appropriate datasets. Low-light datasets such as
LOL~\cite{wei2018retinexnet, yang2021lolv2}, SID~\cite{chen2018sid}, and
MILL~\cite{pilligua2025evaluating} span a narrow range of illuminants, having
been captured under approximately neutral lighting, and provide no illuminant
ground truth against which AWB could be evaluated. As a result, neither
separate enhancement error from color error nor present the challenging
illuminants that make AWB hard in practice.

We address this with \textbf{CHROMA-MILL}. Building on the capture protocol of
MILL~\cite{pilligua2025evaluating}, we capture 50 scenes with a DSLR and
smartphone in RAW with fixed camera parameters, under four calibrated
illuminants---three colored and one white, with measured CIE $xy$
chromaticities---at eleven calibrated intensity levels. Each input is paired
with two intermediate ground truths isolating the individual tasks and a final
ground truth combining them, so the two error sources can be attributed
separately. Because light is additive and the captures are linear RAW and
aligned, we can linearly combine a scene under different illuminants---as in
one-light-at-a-time (OLAT) relighting~\cite{nimeroff1994rerendering,
debevec2000reflectance}---to synthesize it under any convex mixture of those
lights, with exact chromaticity ground truth and no additional capture cost.
This densifies our coverage of the chromaticity plane and brings it closer to the variety of light colors found
in real indoor scenes, each blend acting as a new illuminant of intermediate
chromaticity.

Benchmarking cascades on this data shows that neither ordering is
satisfactory on average: whichever module runs first, hands the second an input outside its
training distribution, and the errors compound at the interface, most
visibly in dark inputs. Joint
correction is better overall, and we introduce \textbf{CTI-Retinexformer} as a
joint baseline to perform
it. It builds on TI-Retinexformer~\cite{pilligua2025evaluating}, which
disentangles illumination intensity from scene content in the latent space, and
extends that disentanglement to color: a new loss supervises an explicit
chromaticity prediction under a uniform-illuminant prior, and the content loss
is redesigned so the scene embedding is invariant to illuminant color as well
as intensity.

\noindent Our contributions are:
\begin{itemize}
\item CHROMA-MILL, the first dataset supporting joint evaluation of low-light
enhancement and white balance: four calibrated illuminants crossed with eleven
calibrated intensity levels over 50 scenes and two cameras, extended with
physically valid illuminant mixtures with exact chromaticity ground truth.
\item The first systematic benchmark of enhancement and white-balance
cascades, in both orderings, quantifying how the two corrections interfere.
\item CTI-Retinexformer, a joint architecture and objective disentangling
intensity, chromaticity, and scene content in the latent space, which
achieves the best overall performance across both cameras.
\end{itemize}

\begin{figure*}[t]
\centering
\setlength{\tabcolsep}{0pt}
\setlength{\fboxsep}{0pt}
\setlength{\fboxrule}{\dsrule}
\setlength{\lineskip}{0pt}\setlength{\lineskiplimit}{0pt}
\renewcommand{\arraystretch}{1}
\setlength{\dsw}{\dimexpr(\linewidth-\dslab-\dsgap-22\dsrule)/11\relax}
\setlength{\dscolw}{\dimexpr\dsw+2\dsrule\relax}

\begin{tabular}{@{}c cccccccccc @{\hspace{\dsgap}} c@{}}
 & \multicolumn{10}{c}{\footnotesize\itshape Low-light inputs}
 & \hd{\itshape $B{=}254$} \\[1.5pt]
 & \hdn{1} & \hdn{2} & \hdn{3} & \hdn{4} & \hdn{5}
 & \hdn{6} & \hdn{7} & \hdn{8} & \hdn{9} & \hdn{10} & \hd{11} \\[1.5pt]
\rowlab{White} & \dscell{\includegraphics[width=\dsw]{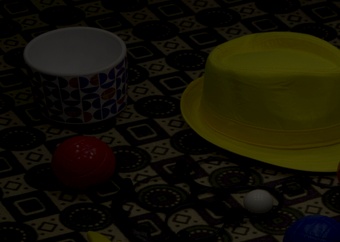}} & \dscell{\includegraphics[width=\dsw]{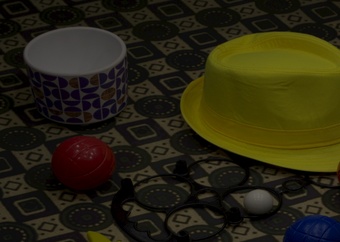}} & \dscell{\includegraphics[width=\dsw]{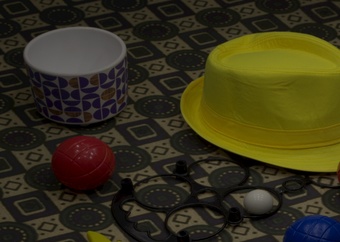}} & \dscell{\includegraphics[width=\dsw]{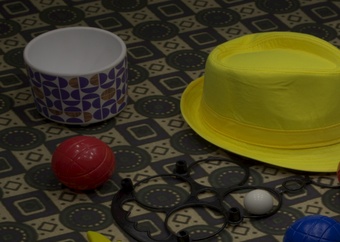}} & \dscell{\includegraphics[width=\dsw]{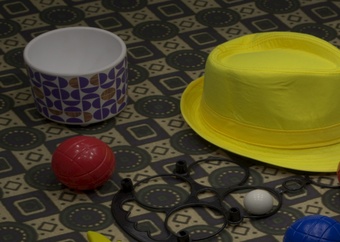}} & \dscell{\includegraphics[width=\dsw]{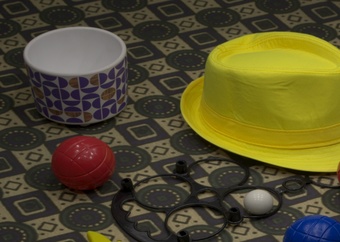}} & \dscell{\includegraphics[width=\dsw]{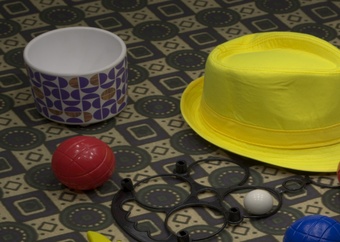}} & \dscell{\includegraphics[width=\dsw]{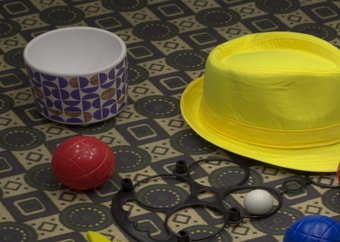}} & \dscell{\includegraphics[width=\dsw]{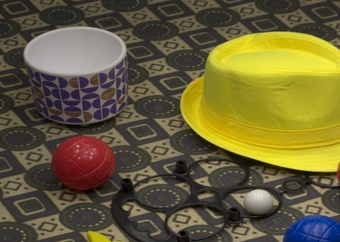}} & \dscell{\includegraphics[width=\dsw]{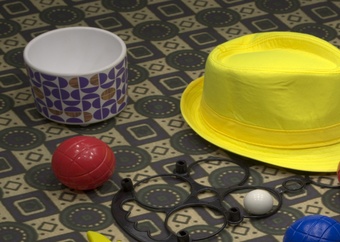}} & \gtcell{\includegraphics[width=\dsw]{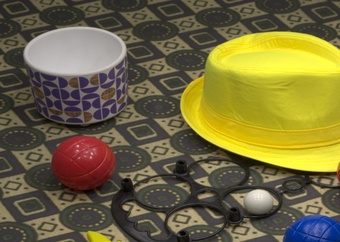}} \\
\rowlab{Red}     & \dscell{\includegraphics[width=\dsw]{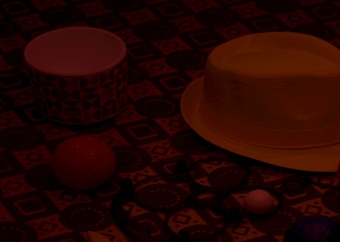}}   & \dscell{\includegraphics[width=\dsw]{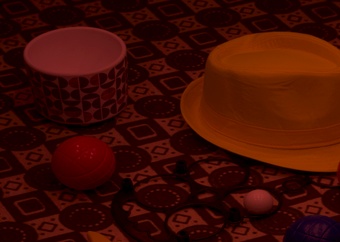}}   & \dscell{\includegraphics[width=\dsw]{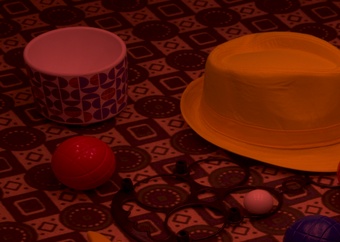}}   & \dscell{\includegraphics[width=\dsw]{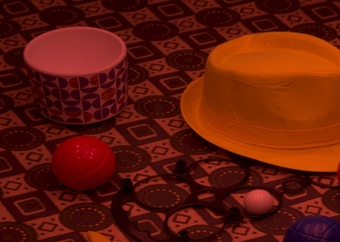}}   & \dscell{\includegraphics[width=\dsw]{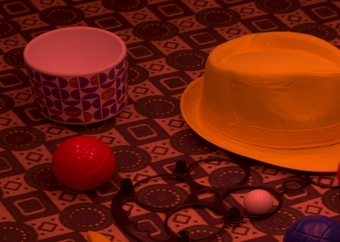}}   & \dscell{\includegraphics[width=\dsw]{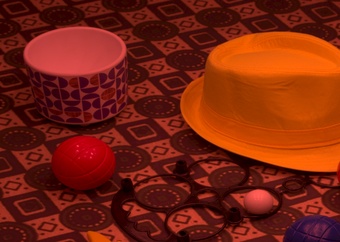}}   & \dscell{\includegraphics[width=\dsw]{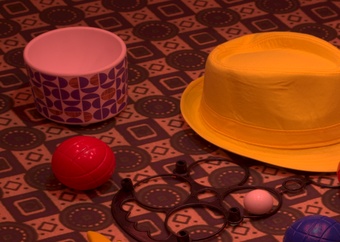}}   & \dscell{\includegraphics[width=\dsw]{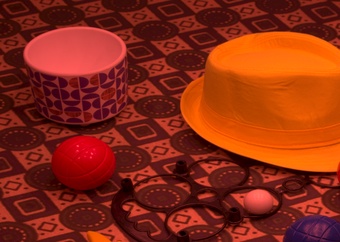}}   & \dscell{\includegraphics[width=\dsw]{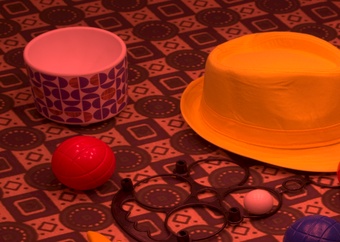}}   & \dscell{\includegraphics[width=\dsw]{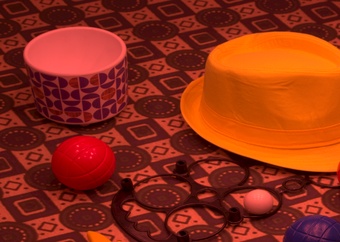}}   & \dscell{\includegraphics[width=\dsw]{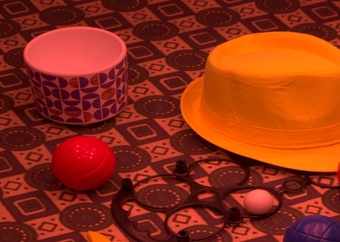}}   \\
\rowlab{Green} & \dscell{\includegraphics[width=\dsw]{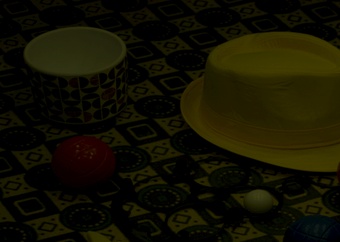}} & \dscell{\includegraphics[width=\dsw]{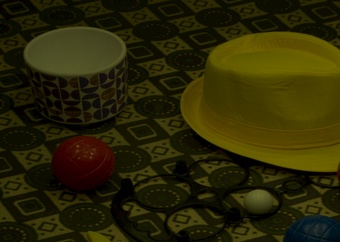}} & \dscell{\includegraphics[width=\dsw]{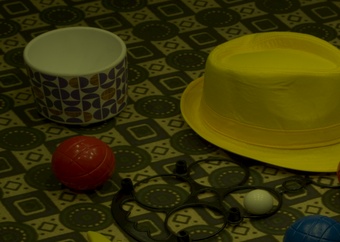}} & \dscell{\includegraphics[width=\dsw]{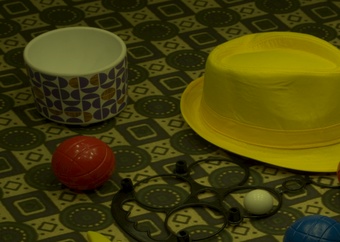}} & \dscell{\includegraphics[width=\dsw]{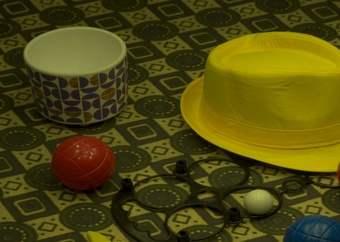}} & \dscell{\includegraphics[width=\dsw]{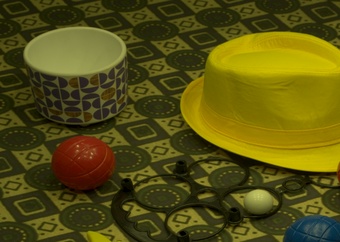}} & \dscell{\includegraphics[width=\dsw]{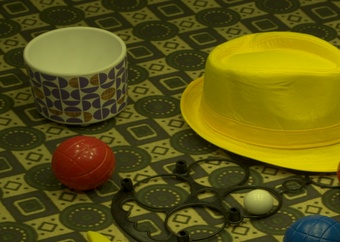}} & \dscell{\includegraphics[width=\dsw]{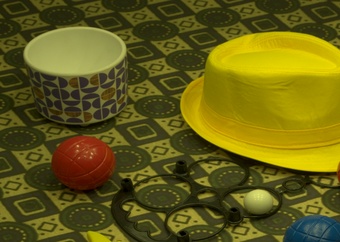}} & \dscell{\includegraphics[width=\dsw]{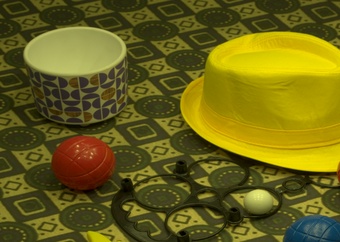}} & \dscell{\includegraphics[width=\dsw]{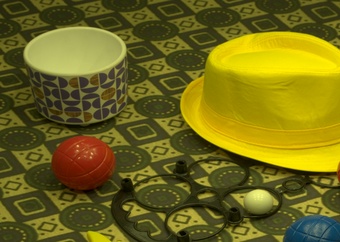}} & \dscell{\includegraphics[width=\dsw]{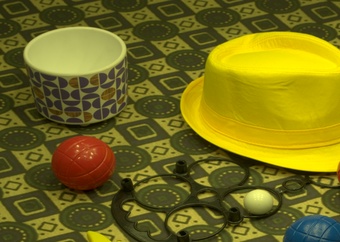}} \\
\rowlab{Blue}   & \dscell{\includegraphics[width=\dsw]{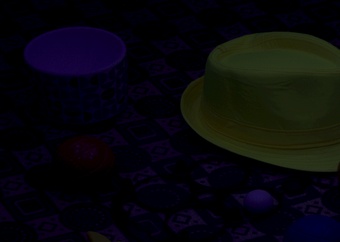}}  & \dscell{\includegraphics[width=\dsw]{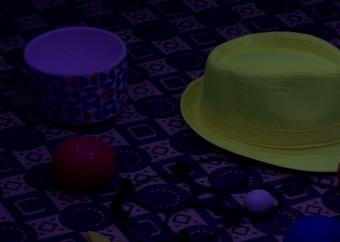}}  & \dscell{\includegraphics[width=\dsw]{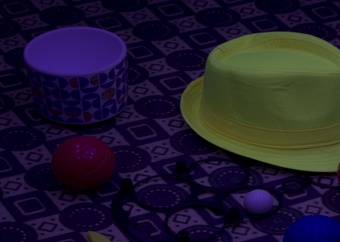}}  & \dscell{\includegraphics[width=\dsw]{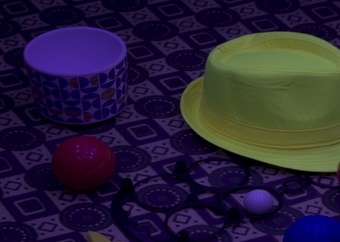}}  & \dscell{\includegraphics[width=\dsw]{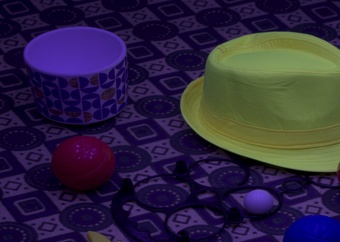}}  & \dscell{\includegraphics[width=\dsw]{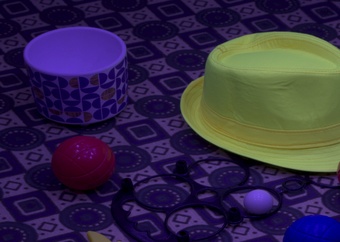}}  & \dscell{\includegraphics[width=\dsw]{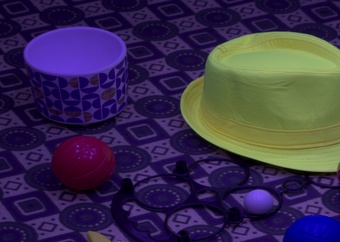}}  & \dscell{\includegraphics[width=\dsw]{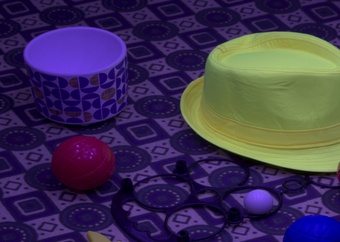}}  & \dscell{\includegraphics[width=\dsw]{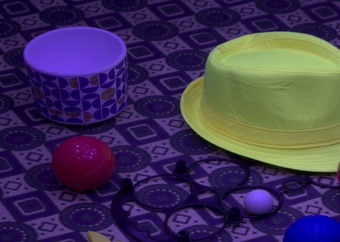}}  & \dscell{\includegraphics[width=\dsw]{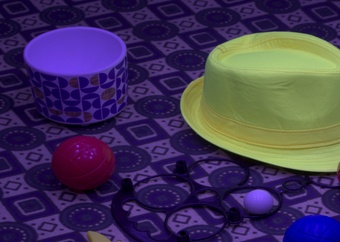}}  & \dscell{\includegraphics[width=\dsw]{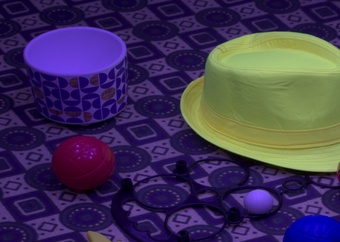}}  \\
\end{tabular}
\caption{One scene of CHROMA-MILL under the four illuminants (white and three colors), shown at the ten low-light levels (Level~1 is the darkest) and at Level~11, each illuminant's maximum-power capture. The white Level~11 capture (\textcolor{cgt}{amber} frame) is the ground truth and is shared by all rows. All cells use the same linear-to-sRGB transform.}
\label{fig:dataset}
\end{figure*}
\section{Related Work}
\label{sec:related}

\subsection{Low-Light Image Enhancement}
Early paired LLIE datasets include LOL~\cite{wei2018retinexnet} and its extension LOLv2~\cite{yang2021lolv2}, captured by varying camera settings, and SID~\cite{chen2018sid}, which introduced RAW-domain capture. Later datasets scaled size and diversity~\cite{liu2021velol, malyugina2023bvi} or exploited multi-exposure sequences~\cite{cai2018sice}. All of them capture each scene under a single low-light condition or simulate brightness change through camera parameters. MILL~\cite{pilligua2025evaluating} addressed this limitation by capturing each scene at eleven calibrated intensity levels with fixed camera settings, revealing that LLIE performance varies substantially across input intensity. However, all these datasets are captured under approximately neutral illumination; none provides color-cast ground truth.

On the method side, classical approaches build on Retinex theory~\cite{landmccann1971lightness}, estimating an illumination map to recover reflectance~\cite{guo2017lime}. RetinexNet~\cite{wei2018retinexnet} introduced learned decomposition, refined by KinD~\cite{zhang2019kind} and KinD++~\cite{zhang2021kindpp}. Reference-free formulations include Zero-DCE~\cite{guo2020zerodce}, SCI~\cite{ma2022sci}, RUAS~\cite{liu2021ruas}, and EnlightenGAN~\cite{jiang2021enlightengan}. Transformer architectures~\cite{zamir2022restormer} now dominate: Retinexformer~\cite{cai2023retinexformer} couples a one-stage Retinex framework with illumination-guided attention, and subsequent work explores frequency-domain modeling~\cite{wang2023fourllie, zhang2025cwnet}, state-space models~\cite{weng2024mamballie}, color-space design~\cite{yan2025cidnet}, geometry-guided normalization~\cite{serrano2025promptnorm}, brightness-mismatch losses~\cite{liao2025gtmean}, and diffusion priors~\cite{yi2023diffretinex, zhou2023pydiff}. Recently, TI-Retinexformer~\cite{pilligua2025evaluating} modified Retinexformer to handle multiple low-light intensity levels. However, like other LLIE methods, it brightens and denoises without estimating or correcting the chromaticity of the scene illuminant.

\subsection{Color Constancy and White Balance}
Statistical illuminant estimation descends from the gray-world assumption~\cite{buchsbaum1980} and the white-patch heuristic~\cite{landmccann1971lightness}, unified in the gray-edge framework~\cite{vandeweijer2007grayedge}. Learned estimators improved accuracy substantially: FC4~\cite{hu2017fc4} pools patch-wise estimates with learned confidence weights, and FFCC~\cite{barron2017ffcc} solves estimation as localization on a torus in the frequency domain. Deep White-Balance Editing~\cite{afifi2020dwb} reframes correction as direct sRGB re-rendering, and~\cite{afifi2019wrongwb} shows that errors baked in by the camera's nonlinear rendering are difficult to undo afterwards. Beyond the single-illuminant assumption, real scenes often have mixed lighting conditions~\cite{afifi2022mixedwb}: LSMI~\cite{kim2021lsmi} provides large-scale captures with per-pixel mixture ratios, AID~\cite{kim2024aid} decomposes illumination with slot attention, and fusion-based correction~\cite{serrano2025fusion} renders multi-illuminant training data at scale. All of these methods presume adequately exposed input, and recent analysis shows deep color constancy models exhibit notable sensitivity to brightness perturbations~\cite{xie2026boosting}. Estimation under severe underexposure remains almost unexplored, with only recent work on nighttime scenes~\cite{lee2026rlawb}. Thus, our benchmark addresses a timely and largely unexplored gap.

\subsection{Joint Restoration}
Joint handling of co-occurring degradations has precedent: LEDNet~\cite{zhou2022lednet} and DarkIR~\cite{feijoo2025darkir} couple low-light enhancement with deblurring, and all-in-one restoration frameworks argue that cascading independently trained modules accumulates errors~\cite{zeng2025allinone}. For enhancement and color, JCRNet~\cite{yu2023joint} balances brightness and color in a multi-stage design, and a zero-shot diffusion prior also performs white balance without task-specific training~\cite{cho2024zeroshot}. Neither trains nor evaluates against measured illuminant ground truth, since, to the authors' knowledge, no dataset providing it currently exists.

In this work, we introduce (i) CHROMA-MILL, a new dataset, and (ii) CTI-Retinexformer, a new model, to overcome the current limitations in the literature. CHROMA-MILL, thanks to its intermediate ground truths (white-illuminant capture at the same brightness, maximum brightness capture with the same color cast as the input) and its final ground truth (white-illuminant capture at maximum brightness), allows training cascaded modules for low-light enhancement and color constancy; however, as will be shown in this work, this cascade training falls short at solving the problem, as cascading models leads to error propagation. This is solved by our new model, CTI-Retinexformer, which introduces a color-constancy loss in the latent space of a state-of-the-art low-light model, enabling joint correction of both tasks.

\section{The CHROMA-MILL Dataset}
\label{sec:dataset}
\begin{table*}[t]
\centering
\caption{Quantitative comparison on the DSLR test set of CHROMA-MILL. Results are averaged over all test images. Every two-stage pipeline is evaluated in both cascade orderings and we report the better one, marking low-light enhancement-first rows with $^{\dagger}$. \colorbox{cbest}{Best}, \colorbox{csecond}{second best}, and \colorbox{cthird}{third best} results are highlighted.}

\label{tab:main}
\footnotesize
\begin{tabular}{l l ccccccc}
\toprule
 & Venue & PSNR$_L$ $\uparrow$ & PSNR$_C$ $\uparrow$ & SSIM $\uparrow$ & LPIPS $\downarrow$ & MAE$_{\angle}$ $\downarrow$ & $\Delta$E$_{76}$ $\downarrow$ & $\Delta$E$_{00}$ $\downarrow$ \\
\midrule
Unprocessed & -- & 18.764 & 17.676 & 0.552 & 0.253 & 23.160 & 22.843 & 16.152 \\
\cdashline{1-9}
\addlinespace[2pt]
TI-Retinexformer~\cite{pilligua2025evaluating} + DWB~\cite{afifi2020dwb} & CVPRF'26 / CVPR'20 & 24.776 & 22.677 & 0.759 & 0.172 & 14.260 & 14.206 & 10.178 \\
TI-Retinexformer~\cite{pilligua2025evaluating} + FC4~\cite{hu2017fc4} & CVPRF'26 / CVPR'17 & 23.784 & 19.546 & 0.653 & 0.280 & 26.035 & 24.476 & 16.652 \\
TI-Retinexformer~\cite{pilligua2025evaluating} + FFCC~\cite{barron2017ffcc}$^{\dagger}$ & CVPRF'26 / CVPR'17 & 23.723 & 19.638 & 0.677 & 0.253 & 26.506 & 24.429 & 16.861 \\
TI-Retinexformer~\cite{pilligua2025evaluating} + RL-AWB~\cite{lee2026rlawb}$^{\dagger}$ & CVPRF'26 / ECCV'26 & 22.915 & 22.034 & 0.685 & 0.183 & 13.618 & 14.603 & 10.792 \\
\addlinespace[2pt]
Retinexformer~\cite{cai2023retinexformer} + DWB~\cite{afifi2020dwb}$^{\dagger}$ & ICCV'23 / CVPR'20 & 22.525 & 19.757 & 0.671 & 0.274 & 18.756 & 18.894 & 13.079 \\
Retinexformer~\cite{cai2023retinexformer} + FC4~\cite{hu2017fc4} & ICCV'23 / CVPR'17 & 23.599 & 19.875 & 0.649 & 0.277 & 26.750 & 23.853 & 16.555 \\
Retinexformer~\cite{cai2023retinexformer} + FFCC~\cite{barron2017ffcc}$^{\dagger}$ & ICCV'23 / CVPR'17 & 23.397 & 19.818 & 0.673 & 0.252 & 27.539 & 24.137 & 16.922 \\
Retinexformer~\cite{cai2023retinexformer} + RL-AWB~\cite{lee2026rlawb} & ICCV'23 / ECCV'26 & 25.666 & 22.885 & 0.702 & 0.195 & 19.149 & 17.665 & 12.041 \\
\addlinespace[2pt]
DarkIR~\cite{feijoo2025darkir} + DWB~\cite{afifi2020dwb} & CVPR'25 / CVPR'20 & \cellcolor{cthird}26.087 & \cellcolor{csecond}23.676 & \cellcolor{cthird}0.783 & 0.165 & 14.738 & 13.787 & 9.838 \\
DarkIR~\cite{feijoo2025darkir} + FC4~\cite{hu2017fc4}$^{\dagger}$ & CVPR'25 / CVPR'17 & 20.548 & 19.239 & 0.622 & 0.272 & 25.926 & 22.302 & 16.755 \\
DarkIR~\cite{feijoo2025darkir} + FFCC~\cite{barron2017ffcc}$^{\dagger}$ & CVPR'25 / CVPR'17 & 23.985 & 20.160 & 0.694 & 0.256 & 27.660 & 24.912 & 17.292 \\
DarkIR~\cite{feijoo2025darkir} + RL-AWB~\cite{lee2026rlawb}$^{\dagger}$ & CVPR'25 / ECCV'26 & 23.486 & 22.470 & 0.706 & 0.178 & \cellcolor{cthird}13.259 & 14.133 & 10.629 \\
\addlinespace[2pt]
MIRNet~\cite{zamir2020mirnet} + DWB~\cite{afifi2020dwb} & ECCV'20 / CVPR'20 & \cellcolor{csecond}26.350 & \cellcolor{cthird}23.663 & \cellcolor{csecond}0.787 & 0.165 & 14.847 & \cellcolor{cthird}13.510 & \cellcolor{csecond}9.697 \\
MIRNet~\cite{zamir2020mirnet} + FC4~\cite{hu2017fc4} & ECCV'20 / CVPR'17 & 23.716 & 19.959 & 0.660 & 0.273 & 26.720 & 23.663 & 16.385 \\
MIRNet~\cite{zamir2020mirnet} + FFCC~\cite{barron2017ffcc}$^{\dagger}$ & ECCV'20 / CVPR'17 & 23.426 & 19.821 & 0.685 & 0.245 & 27.689 & 24.129 & 16.851 \\
MIRNet~\cite{zamir2020mirnet} + RL-AWB~\cite{lee2026rlawb} & ECCV'20 / ECCV'26 & 25.429 & 22.604 & 0.712 & 0.192 & 19.636 & 17.935 & 12.175 \\
\addlinespace[2pt]
PromptNorm~\cite{serrano2025promptnorm} + DWB~\cite{afifi2020dwb}$^{\dagger}$ & CVPRW'25 / CVPR'20 & 22.235 & 20.855 & 0.773 & \cellcolor{cthird}0.157 & 13.495 & 15.173 & 11.221 \\
PromptNorm~\cite{serrano2025promptnorm} + FC4~\cite{hu2017fc4}$^{\dagger}$ & CVPRW'25 / CVPR'17 & 22.886 & 19.909 & 0.678 & 0.251 & 25.027 & 21.959 & 16.531 \\
PromptNorm~\cite{serrano2025promptnorm} + FFCC~\cite{barron2017ffcc}$^{\dagger}$ & CVPRW'25 / CVPR'17 & 25.578 & 19.813 & 0.732 & 0.240 & 26.498 & 25.973 & 17.606 \\
PromptNorm~\cite{serrano2025promptnorm} + RL-AWB~\cite{lee2026rlawb}$^{\dagger}$ & CVPRW'25 / ECCV'26 & 24.864 & 23.429 & 0.753 & \cellcolor{csecond}0.153 & \cellcolor{csecond}11.812 & \cellcolor{csecond}12.618 & \cellcolor{cthird}9.714 \\
\cdashline{1-9}
\addlinespace[2pt]
Ours (CTI-Retinexformer) & -- & \cellcolor{cbest}27.034 & \cellcolor{cbest}24.914 & \cellcolor{cbest}0.809 & \cellcolor{cbest}0.134 & \cellcolor{cbest}11.538 & \cellcolor{cbest}11.144 & \cellcolor{cbest}8.562 \\
\bottomrule
\end{tabular}
\end{table*}
Our goal is to capture a calibrated dataset for training and evaluating joint low-light enhancement at different intensity levels and white-balance correction. We build on the capture protocol of MILL~\cite{pilligua2025evaluating} and extend it along the color axis, collecting over 4,400 RAW captures. All images are captured in a controlled indoor environment, using a platform with
LED programmable lights, a Nikon D5200 DSLR, and a Samsung Galaxy S7
smartphone~\cite{pilligua2025evaluating}. Camera parameters remain fixed
throughout: ISO 100, f/9 and 1/5\,s for the DSLR, f/1.7 and 1/10\,s for the smartphone, following the 95\%-saturation criterion
of~\cite{pilligua2025evaluating}. All images are captured in RAW format and processed in the linear domain. Although LED illuminants can exhibit spectral peaks and do not perfectly reproduce continuous Planckian spectra, they are the preferred technology for fine-grained control of both intensity and
color~\cite{pilligua2025evaluating}.

\paragraph{Colored illuminants and intensity levels.}
We capture each scene under four calibrated illuminants: three colored and one white. The colored illuminants are selected as the maximum red, green, and blue with $x,y$ values of $(0.5235,\,0.3297)$, $(0.2821,\,0.4727)$, and $(0.1968,\,0.1623)$ respectively. The white illuminant measures $(0.3127,\,0.3290)$. All four chromaticities were measured and captured with a Macbeth ColorChecker.

For each illuminant, we capture eleven brightness levels, from minimum to maximum light power linearly based on lux meter readings. The maximum level ($B{=}254$) serves as reference; the remaining ten levels serve as low-light inputs.

Thus, each low-light input is paired with two ``middle'' ground truths: a white-illuminant capture at the same brightness level, which isolates the white-balance task, and a maximum brightness capture with the same color cast, which isolates the low-light enhancement task. The final ground truth is the white-illuminant capture at maximum brightness combining both tasks. Fig.~\ref{fig:dataset} shows the original captured images for a given scene. We capture 50 scenes with both cameras, split into 30 training, 12 validation, and 8 test scenes with no object overlap, following~\cite{pilligua2025evaluating}.
\begin{table*}[t]
\centering
\caption{Quantitative comparison on the phone test set of CHROMA-MILL. Each enhancer is shown with its best-ranking white-balance module; the full grid is in the supplementary. Orderings and highlighting follow Table~\ref{tab:main}.}
\label{tab:main_phone}
\footnotesize
\begin{tabular}{l ccccccc}
\toprule
 & PSNR$_L$ $\uparrow$ & PSNR$_C$ $\uparrow$ & SSIM $\uparrow$ & LPIPS $\downarrow$ & MAE$_{\angle}$ $\downarrow$ & $\Delta$E$_{76}$ $\downarrow$ & $\Delta$E$_{00}$ $\downarrow$ \\
\midrule
Unprocessed & 16.634 & 15.761 & 0.439 & 0.284 & 23.062 & 27.275 & 19.841 \\
\cdashline{1-8}
\addlinespace[2pt]
TI-Retinexformer + DWB$^{\dagger}$ & 22.636 & 21.360 & \cellcolor{cthird}0.638 & \cellcolor{csecond}0.163 & \cellcolor{csecond}11.231 & \cellcolor{cthird}13.379 & 10.162 \\
Retinexformer + DWB$^{\dagger}$ & \cellcolor{cbest}24.421 & \cellcolor{csecond}22.170 & \cellcolor{csecond}0.647 & \cellcolor{cthird}0.166 & \cellcolor{cthird}11.440 & \cellcolor{csecond}12.463 & \cellcolor{csecond}9.327 \\
DarkIR + DWB$^{\dagger}$ & \cellcolor{cthird}23.597 & \cellcolor{cthird}21.545 & 0.627 & 0.167 & 12.249 & 13.395 & \cellcolor{cthird}9.970 \\
MIRNet + DWB$^{\dagger}$ & 23.330 & 21.362 & 0.632 & 0.181 & 12.103 & 13.626 & 10.092 \\
PromptNorm + DWB$^{\dagger}$ & 21.834 & 20.121 & 0.622 & 0.181 & 12.256 & 15.146 & 11.344 \\
\cdashline{1-8}
\addlinespace[2pt]
Ours (CTI-Retinexformer) & \cellcolor{csecond}24.352 & \cellcolor{cbest}22.776 & \cellcolor{cbest}0.649 & \cellcolor{cbest}0.136 & \cellcolor{cbest}10.505 & \cellcolor{cbest}11.387 & \cellcolor{cbest}8.881 \\
\bottomrule
\end{tabular}
\end{table*}
\begin{table*}[t]
\centering
\caption{Performance across input intensity levels on the DSLR test set, from the darkest (Level~1) to the brightest (Level~10). We report PSNR$_L$, PSNR$_C$, MAE$_{\angle}$ and $\Delta$E$_{00}$. Orderings and highlighting follow Table~\ref{tab:main}.}

\label{tab:levels}
\footnotesize
\setlength{\tabcolsep}{3pt}
\resizebox{\textwidth}{!}{%
\begin{tabular}{l cccc cccc cccc cccc}
\toprule
 & \multicolumn{4}{c}{All levels} & \multicolumn{4}{c}{Level 1} & \multicolumn{4}{c}{Level 5} & \multicolumn{4}{c}{Level 10} \\
\cmidrule(lr){2-5}\cmidrule(lr){6-9}\cmidrule(lr){10-13}\cmidrule(lr){14-17}
 & PSNR$_L$ $\uparrow$ & PSNR$_C$ $\uparrow$ & MAE$_{\angle}$ $\downarrow$ & $\Delta$E$_{00}$ $\downarrow$ & PSNR$_L$ $\uparrow$ & PSNR$_C$ $\uparrow$ & MAE$_{\angle}$ $\downarrow$ & $\Delta$E$_{00}$ $\downarrow$ & PSNR$_L$ $\uparrow$ & PSNR$_C$ $\uparrow$ & MAE$_{\angle}$ $\downarrow$ & $\Delta$E$_{00}$ $\downarrow$ & PSNR$_L$ $\uparrow$ & PSNR$_C$ $\uparrow$ & MAE$_{\angle}$ $\downarrow$ & $\Delta$E$_{00}$ $\downarrow$ \\
\midrule
Unprocessed & 18.764 & 17.676 & 23.160 & 16.152 & 11.984 & 11.842 & \cellcolor{cbest}23.318 & 21.523 & 17.726 & 17.317 & 23.478 & 15.750 & 26.631 & 22.830 & 20.891 & 13.061 \\
\cdashline{1-17}
\addlinespace[2pt]
TI-Retinexformer + DWB & 24.776 & 22.677 & \cellcolor{cthird}14.260 & 10.178 & 18.493 & 17.223 & \cellcolor{cthird}25.359 & \cellcolor{cthird}16.058 & 25.520 & 23.384 & 12.411 & 9.284 & 26.125 & 24.160 & 10.413 & 8.387 \\
\addlinespace[2pt]
Retinexformer + RL-AWB & 25.666 & 22.885 & 19.149 & 12.041 & 16.088 & 15.013 & 28.937 & 17.756 & 26.420 & 22.960 & 20.335 & 12.669 & \cellcolor{cbest}30.048 & \cellcolor{cbest}27.509 & \cellcolor{csecond}9.600 & \cellcolor{csecond}7.181 \\
\addlinespace[2pt]
DarkIR + DWB & \cellcolor{cthird}26.087 & \cellcolor{csecond}23.676 & 14.738 & \cellcolor{cthird}9.838 & 19.880 & 18.152 & 30.087 & 18.462 & \cellcolor{csecond}27.337 & \cellcolor{csecond}24.743 & 12.349 & \cellcolor{csecond}8.421 & 27.014 & 24.960 & \cellcolor{cthird}10.328 & \cellcolor{cthird}7.670 \\
\addlinespace[2pt]
MIRNet + DWB & \cellcolor{csecond}26.350 & \cellcolor{cthird}23.663 & 14.847 & \cellcolor{csecond}9.697 & \cellcolor{cthird}20.938 & \cellcolor{cthird}18.176 & 30.293 & 17.160 & \cellcolor{cthird}26.990 & \cellcolor{cthird}24.442 & \cellcolor{cthird}12.306 & \cellcolor{cthird}8.639 & \cellcolor{cthird}27.131 & \cellcolor{cthird}24.977 & 10.358 & 7.718 \\
\addlinespace[2pt]
PromptNorm + DWB$^{\dagger}$ & 22.235 & 20.855 & \cellcolor{csecond}13.495 & 11.221 & \cellcolor{csecond}22.428 & \cellcolor{csecond}19.834 & \cellcolor{csecond}23.328 & \cellcolor{cbest}15.707 & 22.402 & 21.139 & \cellcolor{csecond}12.208 & 10.514 & 22.190 & 21.105 & 10.660 & 10.062 \\
\cdashline{1-17}
\addlinespace[2pt]
Ours (CTI-Retinexformer) & \cellcolor{cbest}27.034 & \cellcolor{cbest}24.914 & \cellcolor{cbest}11.538 & \cellcolor{cbest}8.562 & \cellcolor{cbest}22.432 & \cellcolor{cbest}20.117 & 25.771 & \cellcolor{csecond}16.023 & \cellcolor{cbest}28.216 & \cellcolor{cbest}25.924 & \cellcolor{cbest}9.476 & \cellcolor{cbest}7.362 & \cellcolor{csecond}27.326 & \cellcolor{csecond}25.600 & \cellcolor{cbest}7.906 & \cellcolor{cbest}6.943 \\
\bottomrule
\end{tabular}%
}
\end{table*}
\paragraph{Extending chromatic coverage via superposition.}
The four calibrated illuminants sample the chromaticity plane coarsely. We densify this sampling at no additional capture cost. Because light is additive and our images were captured in the linear RAW domain with static parameters, aligned captures of the same scene under different lighting can be linearly combined to synthesize the same scene under the superposition of those lights.
Concretely, given aligned captures $\{I_c\}$ of a scene at a fixed brightness level, one per calibrated illuminant, we combine them with convex weights in
the linear domain.
\begin{equation}
I_{\text{mix}} = \sum_c \alpha_c I_c, \qquad
\alpha_c \geq 0, \qquad \sum_c \alpha_c = 1 .
\end{equation}
Each weight is a global scalar, so the synthesized image is again lit by a single spatially uniform illuminant, now of intermediate chromaticity.

\begin{figure}[t]
\centering
\setlength{\tabcolsep}{3pt}
\renewcommand{\arraystretch}{1}
\begin{tabular}{@{}cccc@{}}
\includegraphics[width=0.23\linewidth]{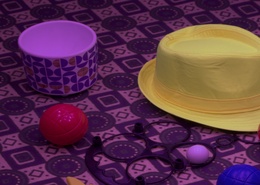} &
\includegraphics[width=0.23\linewidth]{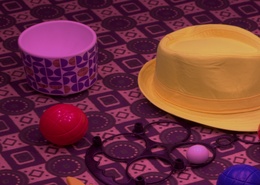} &
\includegraphics[width=0.23\linewidth]{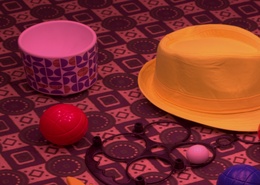} &
\includegraphics[width=0.23\linewidth]{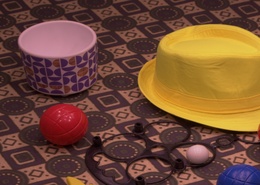} \\
\end{tabular}
\vspace{-1mm}
\caption{Examples of illuminant mixtures: three blends of the blue and red illuminants and one three-illuminant mixture. Same linear-to-sRGB transform as Fig.~\ref{fig:dataset}.}
\label{fig:mixtures}
\end{figure}
The chromaticity of the resulting illuminant follows the same additivity: we recover the CIE XYZ values of each calibrated illuminant from its measured chromaticity and its luminance at full power, combine them with the same weights, and normalize. The result always lies within the convex hull of the calibrated chromaticities. Fig.~\ref{fig:mixtures} shows four examples.

The final dataset contains 14,520 images in training, 5,808 validation images, and 3,872 test images, separated into 30, 12, and 8 scenes respectively at 11 intensity levels.

\section{Method}
\label{sec:method}
\begin{figure*}[t]
\centering
\setlength{\tabcolsep}{2pt}
\setlength{\lineskip}{0pt}\setlength{\lineskiplimit}{0pt}
\renewcommand{\arraystretch}{1}
\setlength{\qlw}{\dimexpr(\linewidth-12\tabcolsep)/7\relax}

\begin{tabular}{@{}ccccccc@{}}
\qlhd{Input} & \qlhd{MIRNet+DWB} & \qlhd{DarkIR+DWB} & \qlhdb{PromptNorm}{+RL-AWB} & \qlhdb{TI-Retinex-}{former+DWB} & \qlhd{\textbf{Ours}} & \qlhd{Ground Truth} \\[1.5pt]
\includegraphics[width=\qlw]{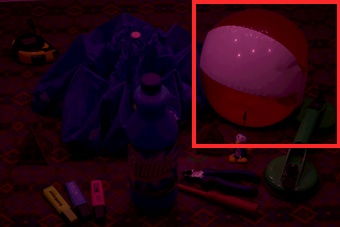} & \includegraphics[width=\qlw]{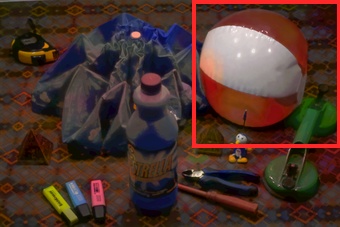} & \includegraphics[width=\qlw]{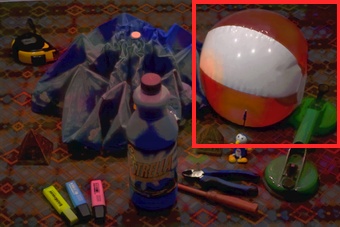} & \includegraphics[width=\qlw]{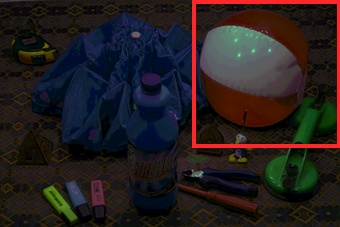} & \includegraphics[width=\qlw]{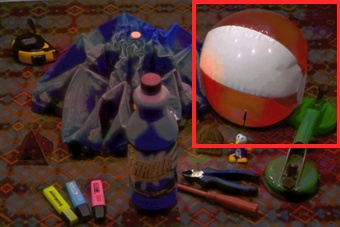} & \includegraphics[width=\qlw]{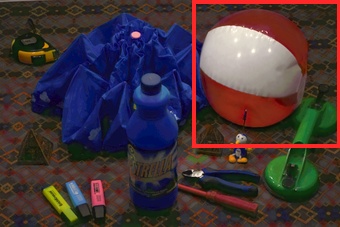} & \includegraphics[width=\qlw]{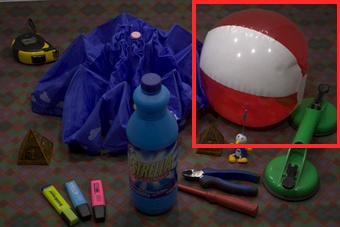} \\
\includegraphics[width=\qlw]{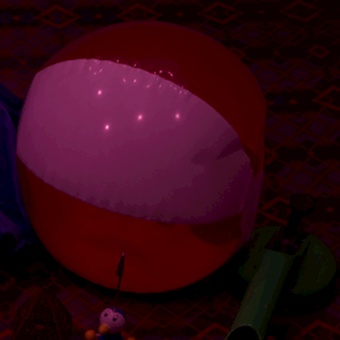} & \includegraphics[width=\qlw]{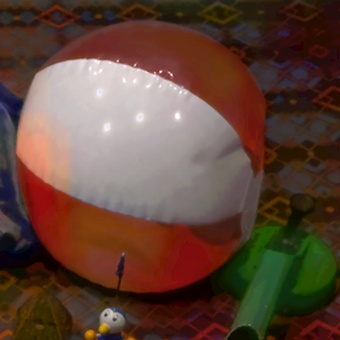} & \includegraphics[width=\qlw]{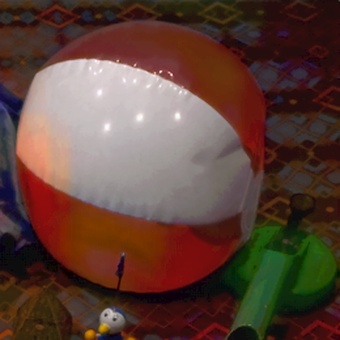} & \includegraphics[width=\qlw]{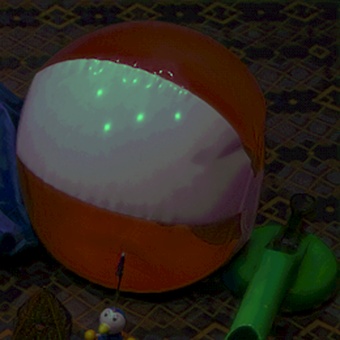} & \includegraphics[width=\qlw]{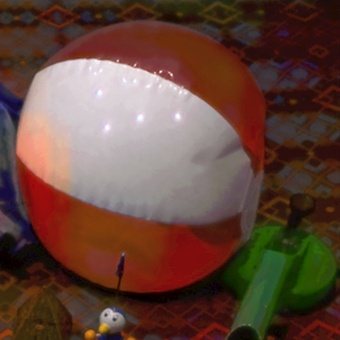} & \includegraphics[width=\qlw]{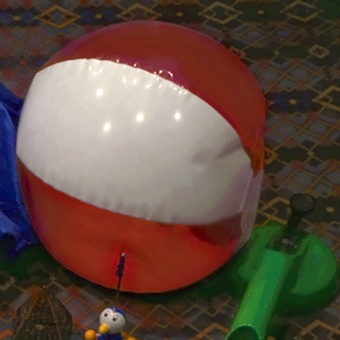} & \includegraphics[width=\qlw]{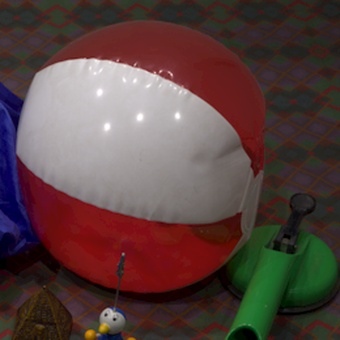} \\[3pt]
\includegraphics[width=\qlw]{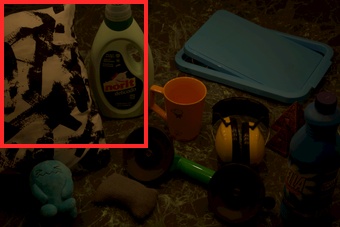} & \includegraphics[width=\qlw]{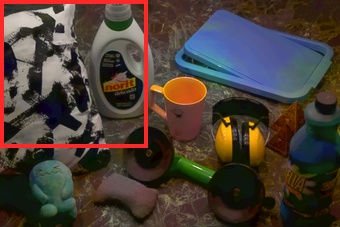} & \includegraphics[width=\qlw]{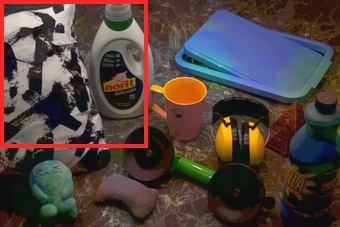} & \includegraphics[width=\qlw]{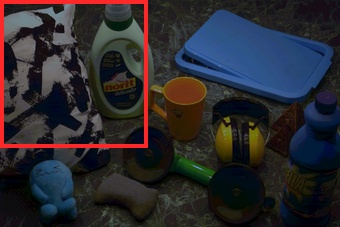} & \includegraphics[width=\qlw]{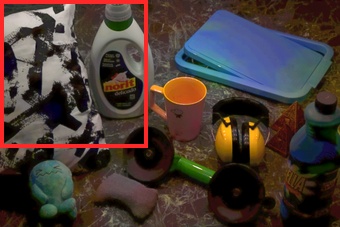} & \includegraphics[width=\qlw]{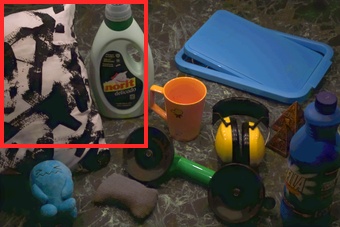} & \includegraphics[width=\qlw]{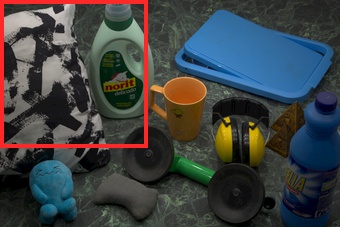} \\
\includegraphics[width=\qlw]{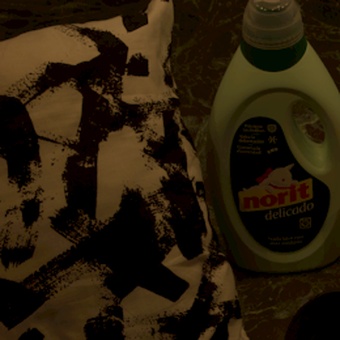} & \includegraphics[width=\qlw]{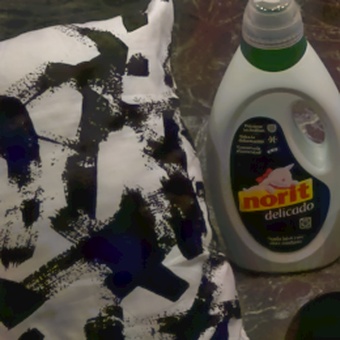} & \includegraphics[width=\qlw]{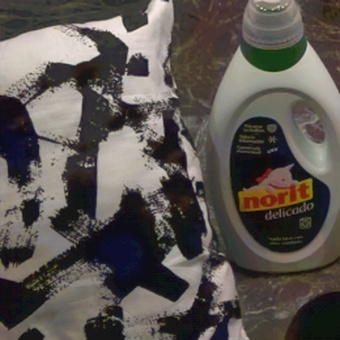} & \includegraphics[width=\qlw]{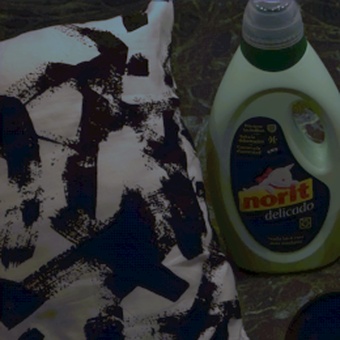} & \includegraphics[width=\qlw]{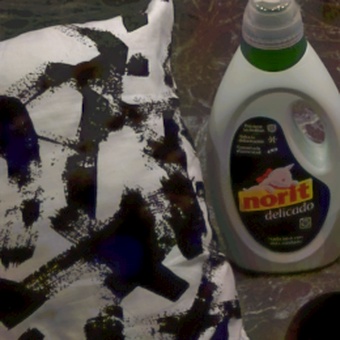} & \includegraphics[width=\qlw]{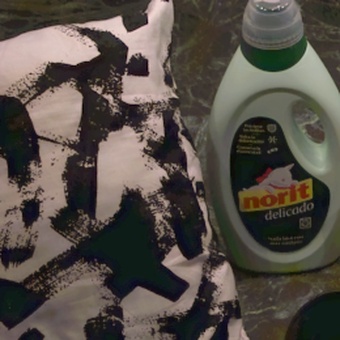} & \includegraphics[width=\qlw]{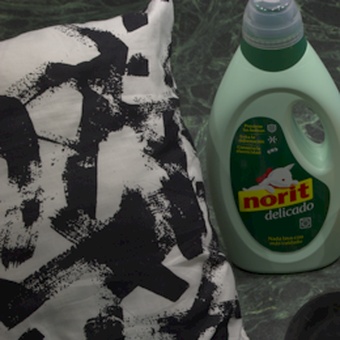} \\[3pt]
\includegraphics[width=\qlw]{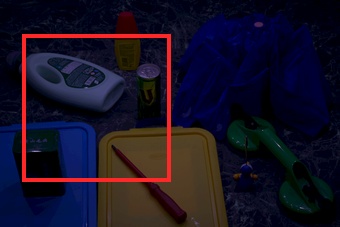} & \includegraphics[width=\qlw]{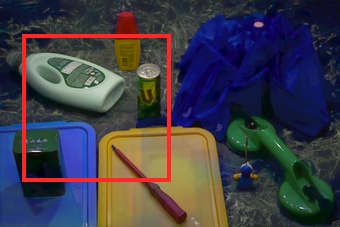} & \includegraphics[width=\qlw]{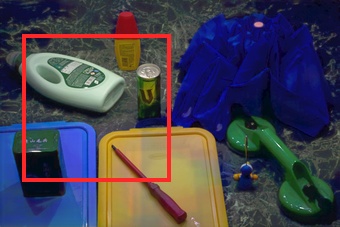} & \includegraphics[width=\qlw]{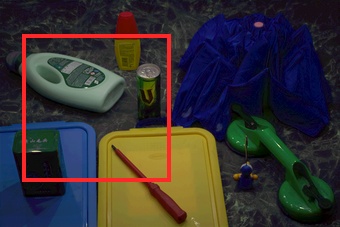} & \includegraphics[width=\qlw]{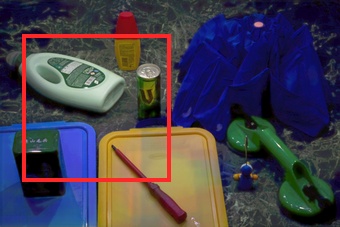} & \includegraphics[width=\qlw]{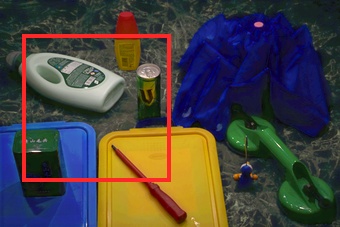} & \includegraphics[width=\qlw]{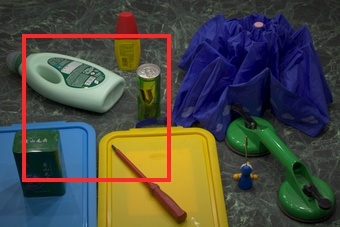} \\
\includegraphics[width=\qlw]{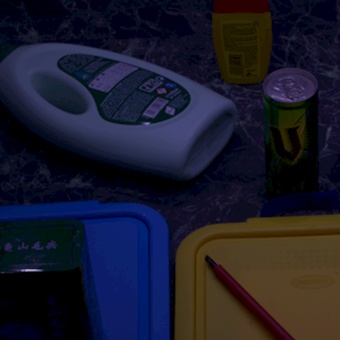} & \includegraphics[width=\qlw]{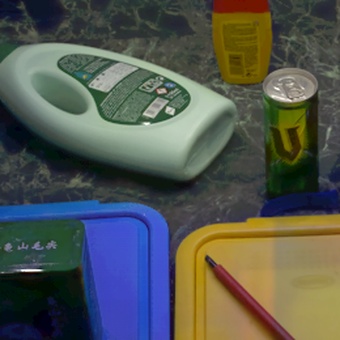} & \includegraphics[width=\qlw]{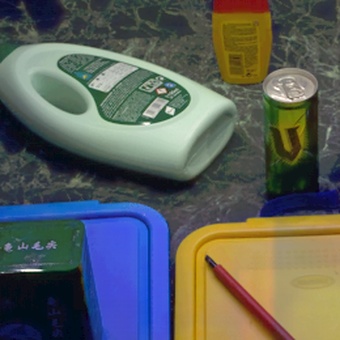} & \includegraphics[width=\qlw]{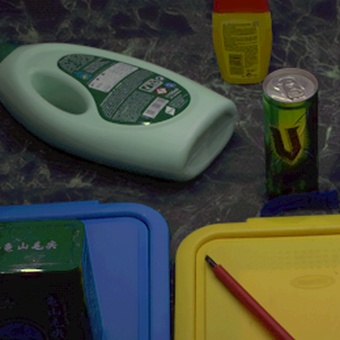} & \includegraphics[width=\qlw]{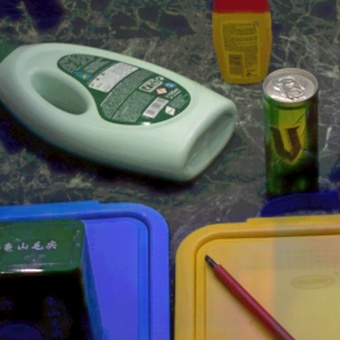} & \includegraphics[width=\qlw]{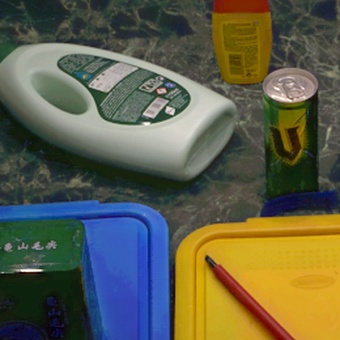} & \includegraphics[width=\qlw]{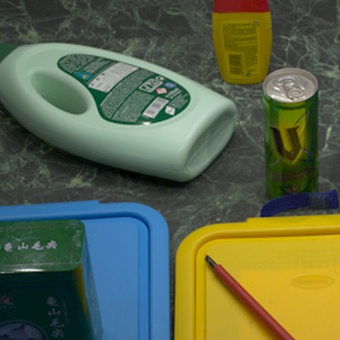} \\
\end{tabular}
\vspace{-2mm}
\caption{Visual comparison on CHROMA-MILL. From left to right: input, MIRNet+DWB, DarkIR+DWB, PromptNorm+RL-AWB, TI-Retinexformer+DWB, ours, and ground truth. We show three examples under different illuminant colors, with a zoomed-in region below each. The baselines are the four best-performing pipelines of Table~\ref{tab:main}. Same linear-to-sRGB transform as Fig.~\ref{fig:dataset}.}
\label{fig:qualitative}
\end{figure*}

We propose CTI-Retinexformer, a method that disentangles intensity, color, and scene content in the latent space. The backbone architecture is Retinexformer~\cite{cai2023retinexformer}, an architecture comprising an illumination estimator and a U-shaped denoiser built from illumination-guided attention blocks, with base width $C=40$ and channels doubling at each of two downsampling stages, giving $4C=160$ channels in the latent space.

In TI-Retinexformer~\cite{pilligua2025evaluating}, the authors partitioned these latent features of Retinexformer into an intensity channel and a scene embedding. Our model goes one step further by extending the latent disentanglement strategy from intensity to color: we retain the intensity channel $Z_I$ (1 channel), modify the scene embedding $Z_{s}$ to allocate for color variations (157 channels), and propose a new chromaticity prediction region $Z_{ch}$ (2 channels). Each component is supervised by a dedicated loss. In what follows, we explain the different disentanglement losses.

\paragraph{Intensity prediction loss.} Following~\cite{pilligua2025evaluating}, the first channel predicts the normalized intensity $i_{in} \in [0,1]$ of the input, given by its light-power level, $i_{in} = B/254$. We penalize the squared error of its spatial mean:
\begin{equation}
\mathcal{L}_{i} = \left( \overline{Z_I} - i_{in} \right)^2,
\end{equation}
where $\overline{\cdot}$ denotes the spatial mean.

\paragraph{Chromaticity losses.} We propose new chromaticity losses to focus on the color constancy task. In detail, our goal is that the second and third channels of the latent space predict the CIE $xy$ chromaticity $c_{in}$ of the scene illuminant:
\begin{equation}
\mathcal{L}_{c} = \left\| \overline{Z_{ch}} - c_{in} \right\|_2^2 .
\end{equation}

Moreover, since each training image is lit by a spatially uniform (single or uniformly blended) illuminant, we additionally penalize the spatial variance of the prediction to force the encoding to output a uniform-illuminant prior.
\begin{equation}
\mathcal{L}_{u} = \mathrm{Var}(Z_{ch}).
\end{equation}

\paragraph{Scene content loss.} In~\cite{pilligua2025evaluating}, they proposed that the remaining channels should encode the content of the scene independently of illumination. In our case, we extend this scene content loss to color. For that, we consider a triplet margin loss~\cite{schroff2015facenet} in which the positive is sampled from the same scene as the anchor under a different illumination condition ---a different brightness level, a different illuminant color, or both--- and the negative from a different scene. Anchor and positive share identical crop coordinates to enforce spatial correspondence.
\begin{equation}
\mathcal{L}_{tr} = \max\!\left( \|Z_s^{q} - Z_s^{p}\|_2 - \|Z_s^{q} - Z_s^{n}\|_2 + m,\; 0 \right).
\end{equation}
where $Z_s^{q}$, $Z_s^{p}$, and $Z_s^{n}$ $\in$ $\mathbb{R}^{\frac{H}{4}\times \frac{W}{4} \times (4C-3)}$ correspond to the anchor, positive and negative latent features after the bottleneck attention block, and $m{=}1$; distances are taken along each feature-map element and later averaged.

\paragraph{Reconstruction loss.} We also add a reconstruction loss $\mathcal{L}_{re}$ defined as the mean L1 distance between the network output and the ground truth image.
 \paragraph{Final loss.}
 The complete objective loss is the unweighted sum of all the aforementioned ones:
\begin{equation}
\mathcal{L} = \mathcal{L}_{re} + \mathcal{L}_{i} + \mathcal{L}_{c} + \mathcal{L}_{u} + \mathcal{L}_{tr}.
\end{equation}

\section{Experiments}
\label{sec:experiments}
\begin{figure*}[t]
\centering
\setlength{\tabcolsep}{2pt}
\setlength{\lineskip}{0pt}\setlength{\lineskiplimit}{0pt}
\renewcommand{\arraystretch}{1}
\setlength{\qlw}{\dimexpr(\linewidth-10\tabcolsep-1pt)/6\relax}

\begin{tabular}{@{}cccccc@{}}
\qlhd{Input} & \qlhd{MIRNet+DWB} & \qlhd{DarkIR+DWB} & \qlhdb{PromptNorm}{+RL-AWB} & \qlhdb{TI-Retinex-}{former+DWB} & \qlhd{\textbf{Ours}} \\[1.5pt]
\includegraphics[width=\qlw]{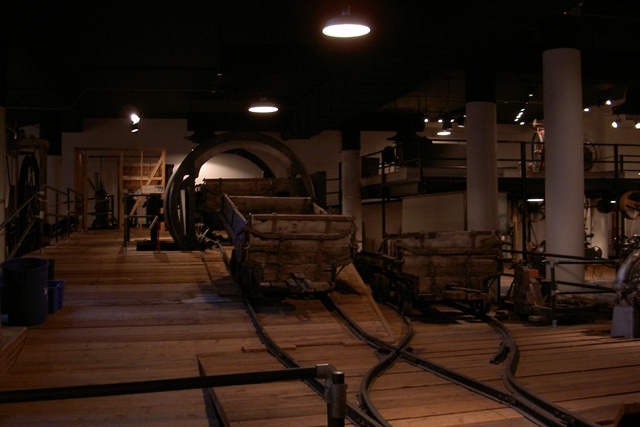} & \includegraphics[width=\qlw]{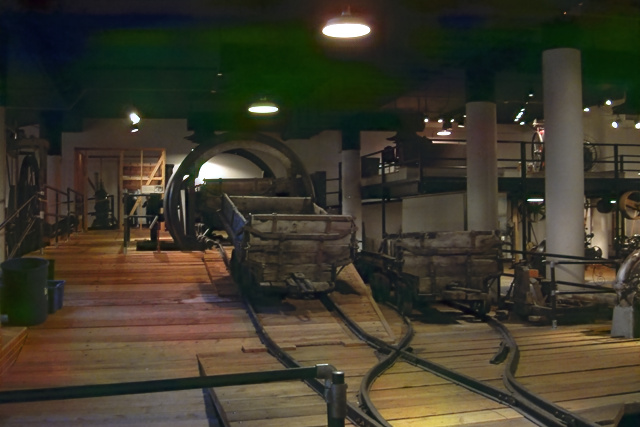} & \includegraphics[width=\qlw]{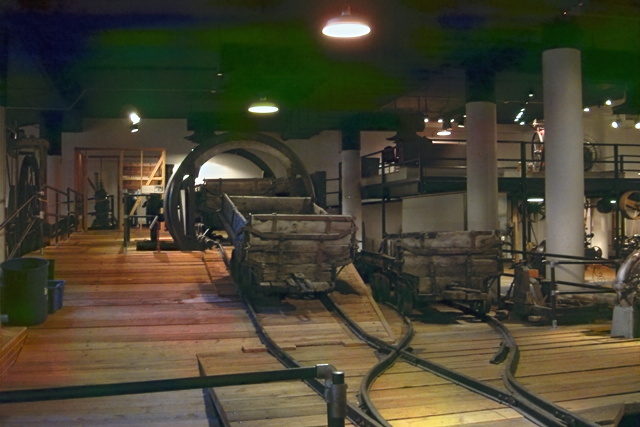} & \includegraphics[width=\qlw]{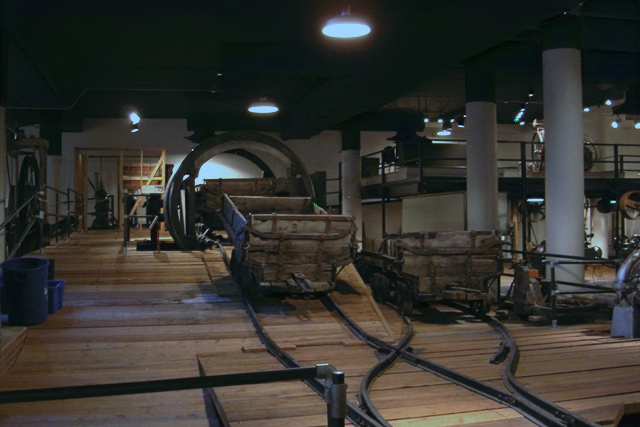} & \includegraphics[width=\qlw]{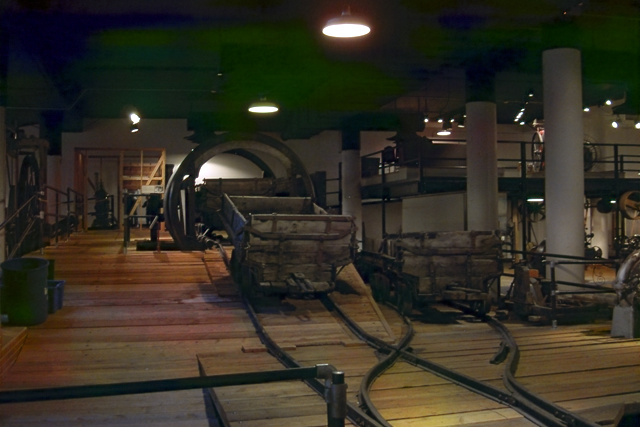} & \includegraphics[width=\qlw]{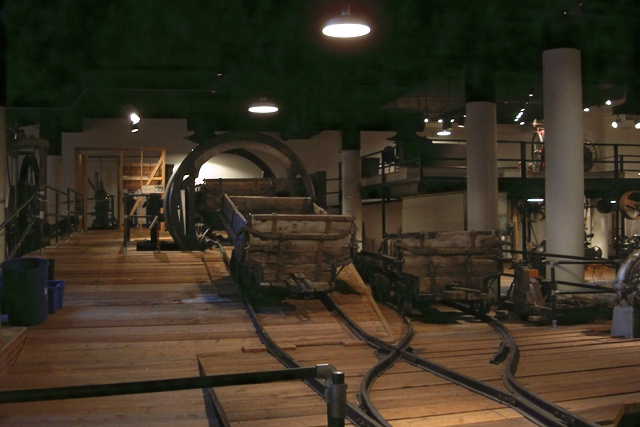} \\
\includegraphics[width=\qlw]{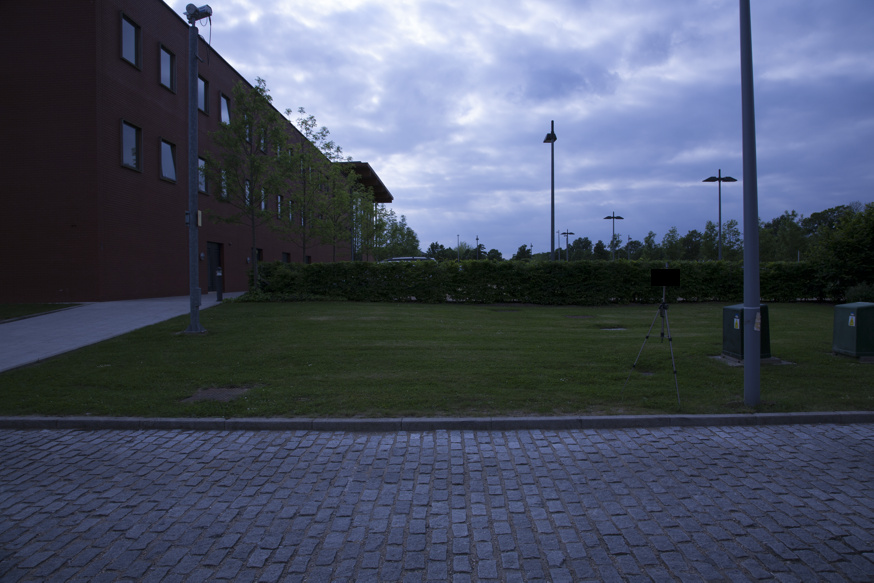} & \includegraphics[width=\qlw]{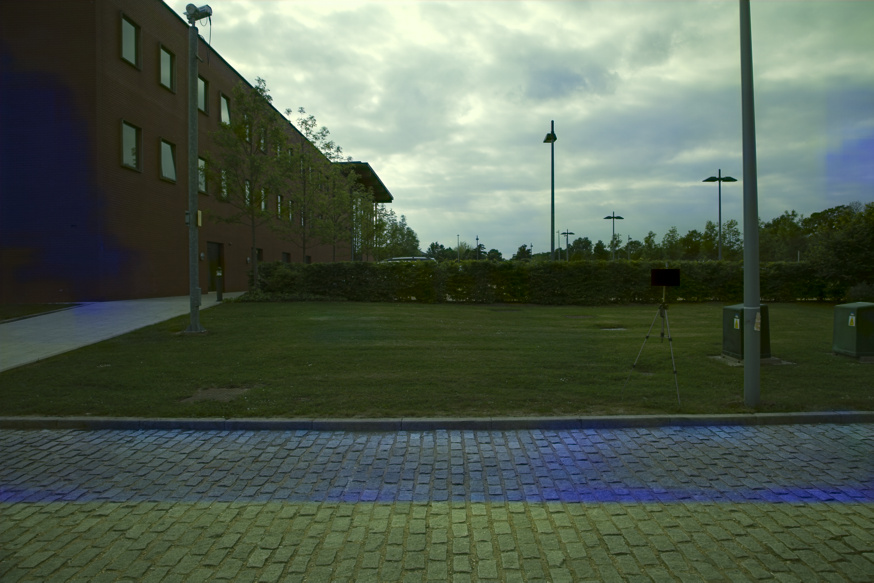} & \includegraphics[width=\qlw]{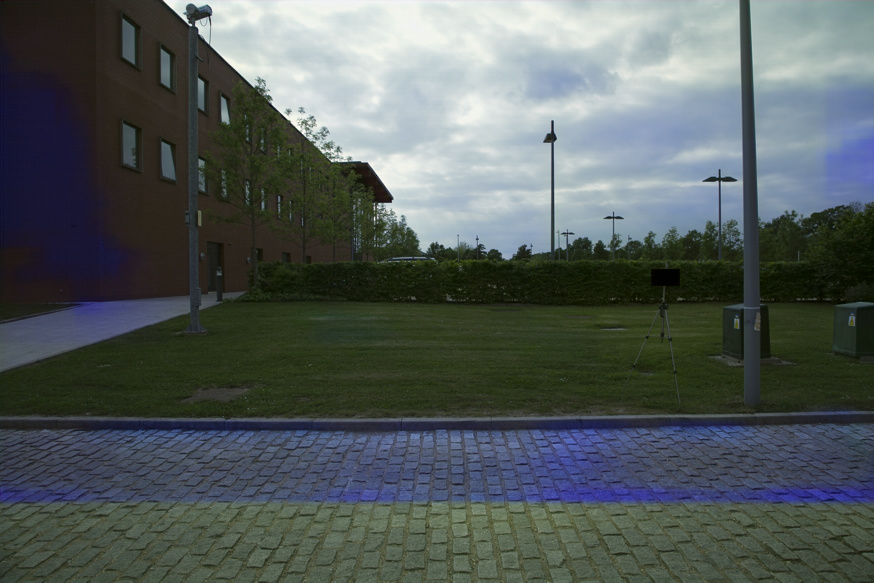} & \includegraphics[width=\qlw]{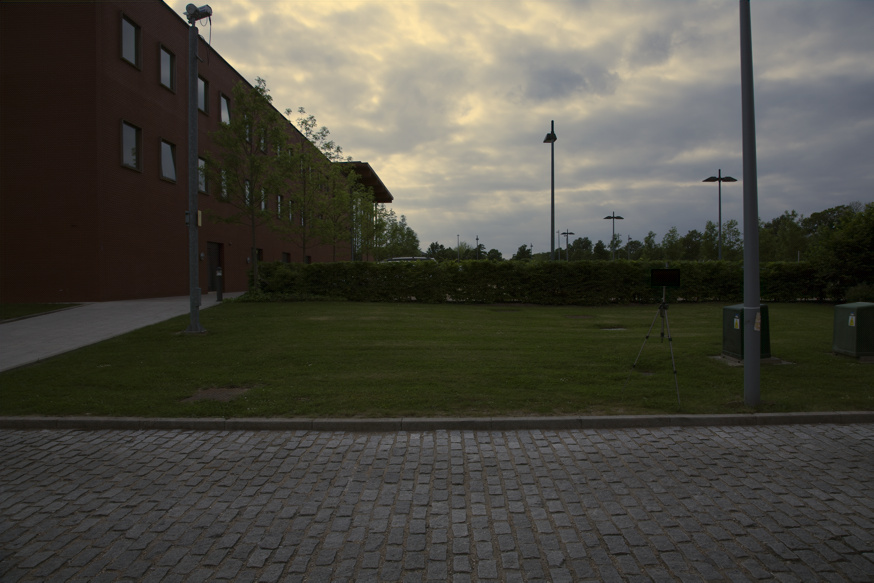} & \includegraphics[width=\qlw]{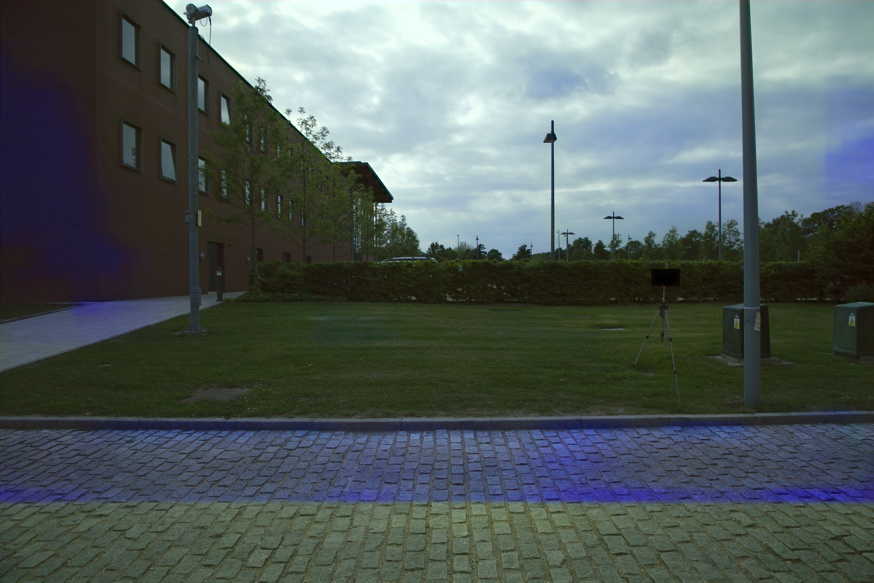} & \includegraphics[width=\qlw]{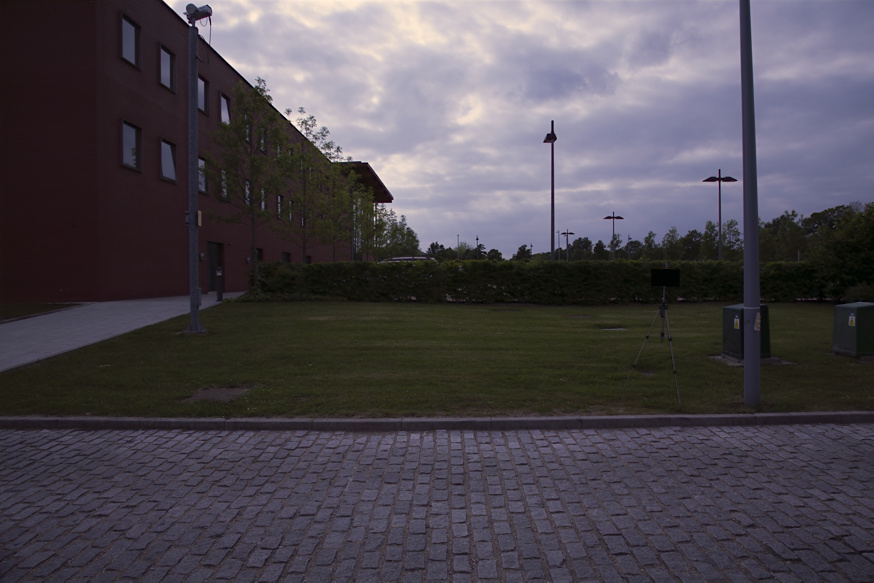} \\
\includegraphics[width=\qlw]{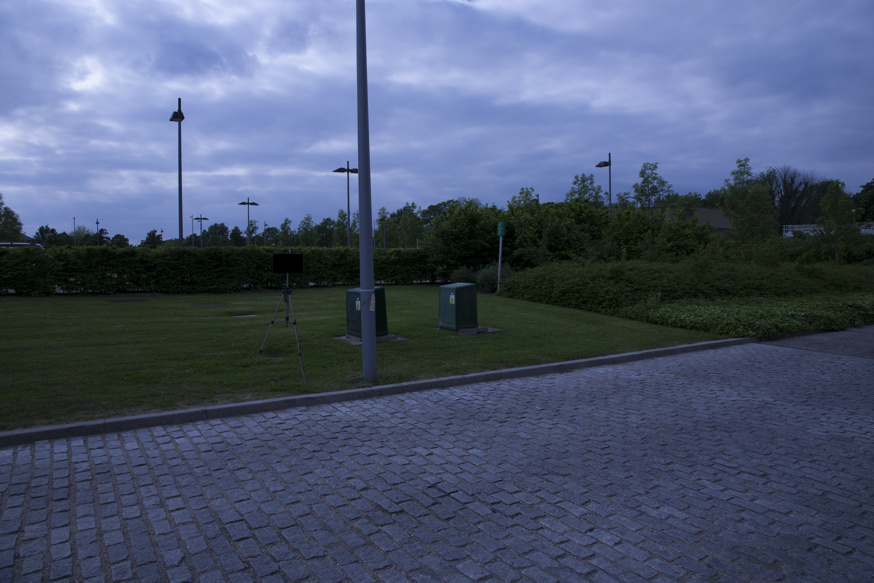} & \includegraphics[width=\qlw]{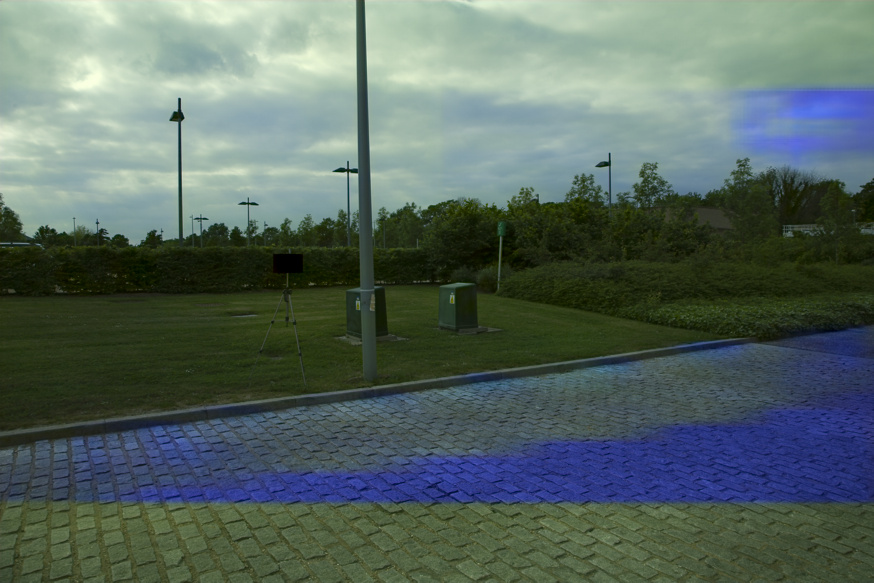} & \includegraphics[width=\qlw]{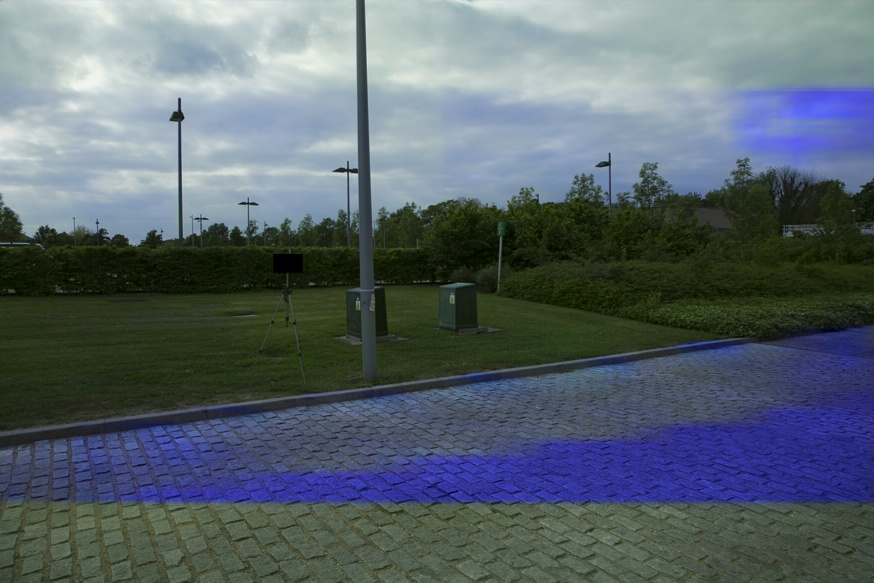} & \includegraphics[width=\qlw]{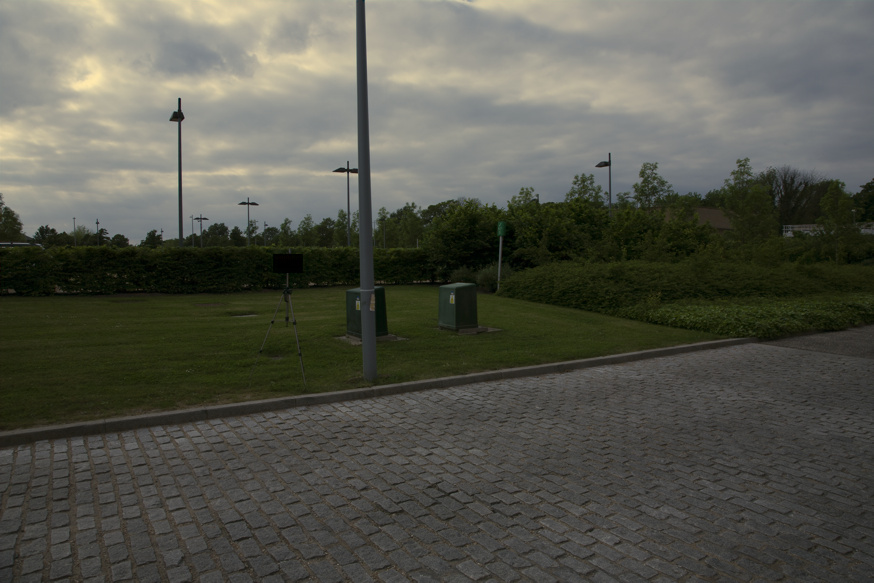} & \includegraphics[width=\qlw]{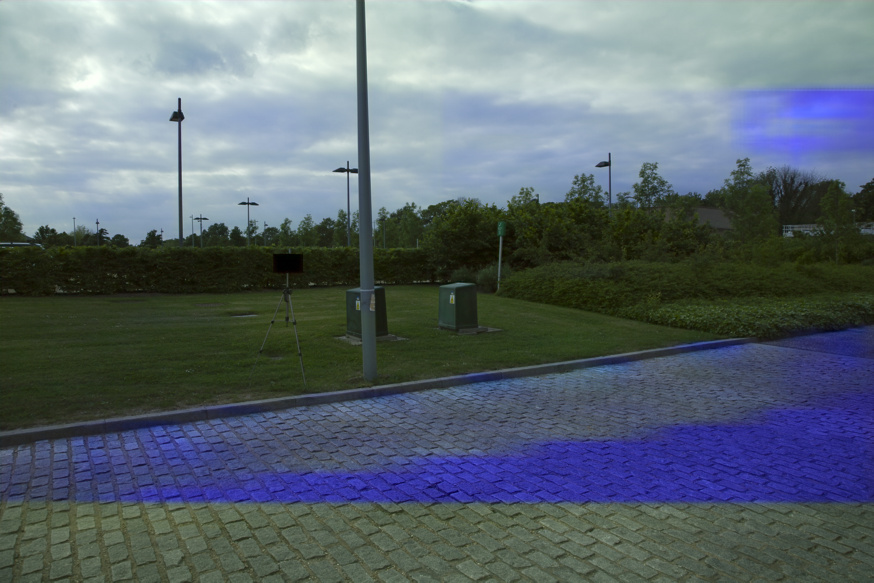} & \includegraphics[width=\qlw]{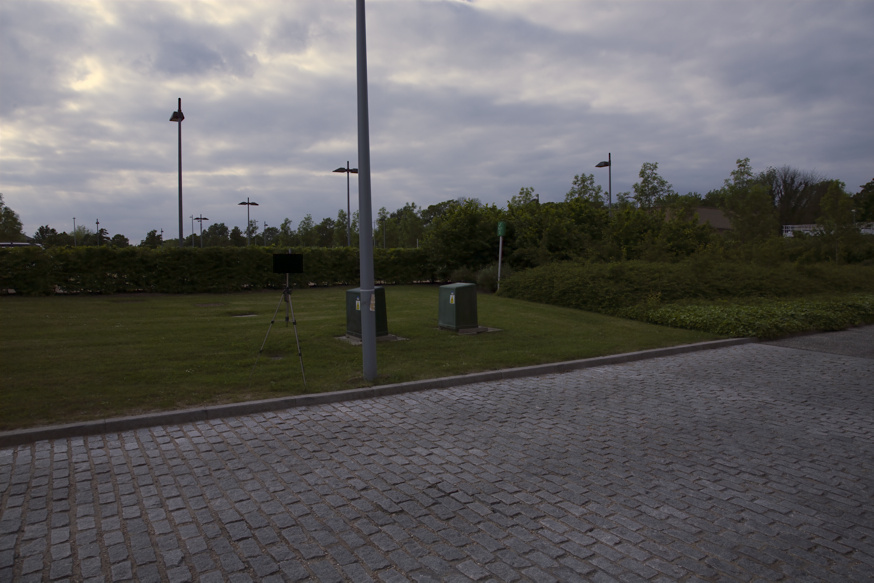} \\
\includegraphics[width=\qlw]{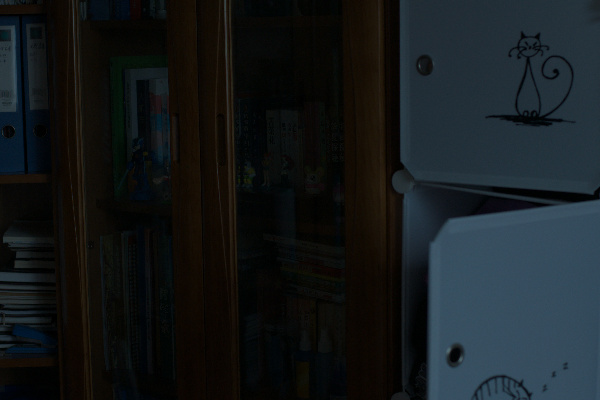} & \includegraphics[width=\qlw]{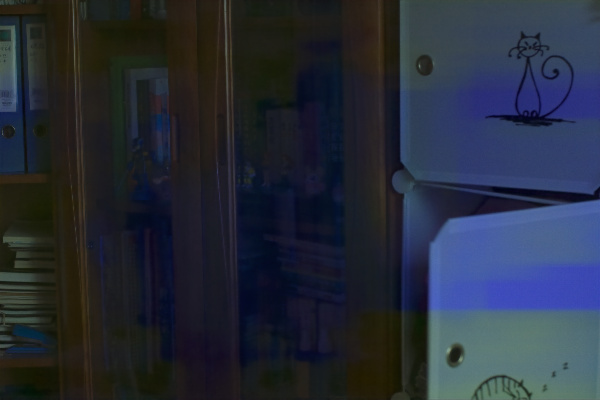} & \includegraphics[width=\qlw]{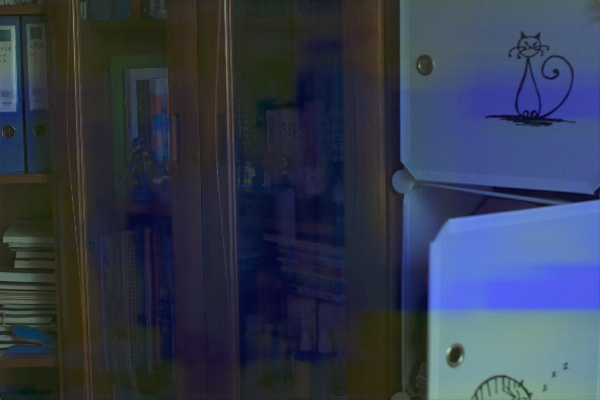} & \includegraphics[width=\qlw]{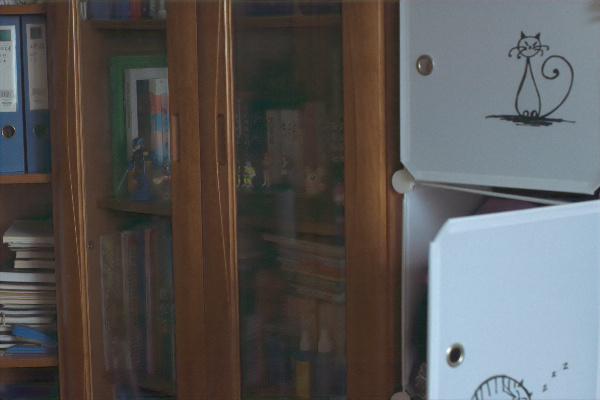} & \includegraphics[width=\qlw]{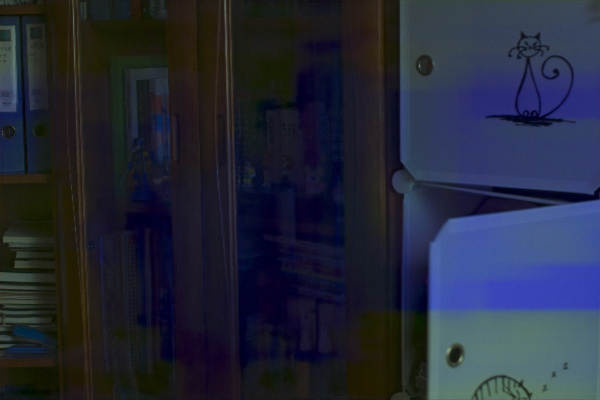} & \includegraphics[width=\qlw]{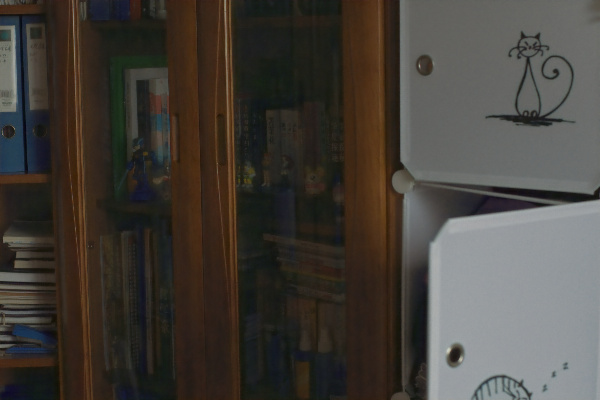} \\
\end{tabular}
\caption{Generalization to other datasets. Rows, top to bottom: one image from DICM~\cite{lee2012dicm}, two from the Rendered WB dataset~\cite{afifi2019wrongwb}, and one from LOLv1~\cite{wei2018retinexnet}. Columns follow Fig.~\ref{fig:qualitative}: input, the four best-performing pipelines of Table~\ref{tab:main}, and ours. No ground truth is available for these datasets, so no metrics are reported. All models are trained on CHROMA-MILL only, with no fine-tuning. Same linear-to-sRGB transform as Fig.~\ref{fig:dataset}.}
\vspace{-4mm}
\label{fig:external}
\end{figure*}

\subsection{Benchmark on CHROMA-MILL}
\label{sec:benchmark}
We compare our approach against two-stage pipelines pairing a low-light enhancer with a white-balance module, and against the unprocessed input. The enhancers are Retinexformer~\cite{cai2023retinexformer}, TI-Retinexformer~\cite{pilligua2025evaluating}, DarkIR~\cite{feijoo2025darkir}, MIRNet~\cite{zamir2020mirnet}, and PromptNorm~\cite{serrano2025promptnorm}; the white-balance modules are Deep White-Balance Editing~\cite{afifi2020dwb}, FC4~\cite{hu2017fc4}, FFCC~\cite{barron2017ffcc}, and RL-AWB~\cite{lee2026rlawb}. All models are fine-tuned on the respective training set of CHROMA-MILL, using the intermediate ground truths. We report PSNR on luminance (PSNR$_L$) and RGB (PSNR$_C$)~\cite{pilligua2025evaluating}, SSIM~\cite{wang2004ssim}, LPIPS~\cite{zhang2018lpips}, mean recovery angular error~\cite{hordley2006reevaluation}, $\Delta E_{76}$, and $\Delta E_{2000}$~\cite{sharma2005ciede2000}. Every two-stage pipeline is evaluated in both cascade orderings ---white balance + enhance or enhance + white balance. Table~\ref{tab:main} shows the results for the Nikon camera, reporting the ordering that ranks higher across the seven metrics; the complete evaluation with both orderings listed separately is in the supplementary.

Our model leads on every metric in Table~\ref{tab:main}, and the margin is widest on color. Against MIRNet+DWB, the strongest cascade on fidelity, it gains $0.68$\,dB PSNR$_L$ and lowers $\Delta E_{00}$ by $1.14$ and $\Delta E_{76}$ by $2.37$; against PromptNorm+RL-AWB, the strongest on color, it gains $2.17$\,dB PSNR$_L$ and lowers $\Delta E_{00}$ by $1.15$ and $\Delta E_{76}$ by $1.47$. Fidelity and color improve together, and neither is traded for the other; Section~\ref{sec:ablations} examines the contribution of each auxiliary loss term.

Two-stage pipelines suffer from error compounding, and no cascade ordering avoids it. A stronger enhancer reduces the fidelity gap but not the color gap: DarkIR~\cite{feijoo2025darkir} and MIRNet~\cite{zamir2020mirnet}, both cascaded with Deep White-Balance Editing~\cite{afifi2020dwb}, come within 1 dB of our PSNR$_L$ while remaining more than one $\Delta E_{00}$ unit behind. PromptNorm cascaded with RL-AWB is the strongest competitor on color metrics, but it still trails on all metrics and gives up $2.2$\,dB of PSNR$_L$. This behavior is expected, since the residuals of the first stage are out-of-distribution inputs for the second. In contrast, a single network sharing features across both tasks has no such interface. The cascade ordering does not compensate for the decomposition on average. Which ordering performs better depends on the pipeline: white-balance-first for the DWB cascades built on DarkIR and MIRNet, and enhancement-first for all the PromptNorm pipelines. The difference between the orderings can exceed $3$\,dB of PSNR$_L$, so it must be tuned for each pair of methods. Even so, averaged over all levels, the better of the two orderings never reaches our jointly trained model.

Table~\ref{tab:main_phone} provides the results for the Smartphone test set, with all the models retrained on the phone training split. Only the best combination for each low-light enhancer is shown; the full table is in the supplementary. Our method obtains the best results on every metric except PSNR$_L$ where Retinexformer cascaded with Deep White-Balance Editing is $0.07$\,dB higher while being worse on the six remaining metrics. The margin on color is even larger than on the Nikon split. The absolute values are lower for all the methods due to the noisier sensor.

Table~\ref{tab:levels} reports performance across input intensity levels for the Nikon camera. Only the best results from Table \ref{tab:main} are presented. Our method obtains the best results in 12 out of 16 cases, being second in 3 other ones. The advantage is not uniform across levels: at the brightest inputs (Levels~7--10), Retinexformer+RL-AWB leads on PSNR$_L$ by up to $2.7$\,dB, while ours still leads on $\Delta E_{00}$. At high SNR, enhancement is close to the identity and the white-balance module receives a well-exposed input, so little error compounds at the interface.

\subsection{Qualitative Results}
\label{sec:qualitative}

Figure~\ref{fig:qualitative} shows three examples at Levels~2 and~3, under two synthesized blends and the pure blue primary, against the four strongest baselines of Table~\ref{tab:main}. In each case, CTI-Retinexformer leads all four on PSNR$_L$, PSNR$_C$, and $\Delta E_{00}$. The failure modes depend more on the illuminant than on the enhancer. The three DWB cascades \emph{overcorrect} the two blends, carrying whites past neutral, and \emph{undercorrect} the blue primary, leaving a deep blue cast. PromptNorm+RL-AWB shifts the cast rather than removing it and reconstructs the scene far more darkly, more than $5$\,dB below the DWB cascades on PSNR$_L$ in both blends. Both failure modes are consistent with the error compounding of Sec.~\ref{sec:benchmark}. Our joint model stays closest to the ground truth, though the blue primary remains hard for it too, with a residual cast and visible grain.
\vspace{-5mm}
\paragraph{Generalization to other datasets.}
\label{sec:external}
Figure~\ref{fig:external} applies the models of Table~\ref{tab:main} to DICM~\cite{lee2012dicm}, Rendered WB~\cite{afifi2019wrongwb} and LOLv1~\cite{wei2018retinexnet} images without fine-tuning. On the Rendered WB scenes the cascades reproduce their failure modes under domain shift: MIRNet+DWB overshoots into a yellow-green cast, while DarkIR+DWB and TI-Retinexformer+DWB leave much of the input's blue tint. CTI-Retinexformer removes the cast most completely while preserving the warm tungsten character of the DICM. On the last example PromptNorm+RL-AWB gives a brighter exposure, but ours still avoids the strong blue cast of the other cascades.
\subsection{Ablations}

\begin{table}[t]
\centering
\caption{Ablation of the loss terms on the DSLR test set. Each row enables a different subset of the auxiliary losses; the reconstruction loss $\mathcal{L}_{re}$ is always active and every row is trained jointly to the final ground truth, from scratch, with the same training configuration.}

\label{tab:ablation}
\footnotesize
\setlength{\tabcolsep}{2pt}
\resizebox{\linewidth}{!}{%
\begin{tabular}{l cccc cccc}
\toprule
Model & $\mathcal{L}_{i}$ & $\mathcal{L}_{tr}$ & $\mathcal{L}_{c}$ & $\mathcal{L}_{u}$ & PSNR$_L$ $\uparrow$ & PSNR$_C$ $\uparrow$ & MAE$_{\angle}$ $\downarrow$ & $\Delta$E$_{76}$ $\downarrow$ \\
\midrule
Rec.\ only & & & & & 24.885 & 23.522 & \textbf{11.255} & 11.329 \\
I  & \checkmark &            &            &            & 26.408 & 24.460 & 11.641 & 11.309 \\
S  &            & \checkmark &            &            & 25.632 & 23.735 & 12.306 & 12.345 \\
TI & \checkmark & \checkmark &            &            & 26.208 & 24.374 & 11.913 & 11.547 \\
CTI w/o $\mathcal{L}_{u}$ & \checkmark & \checkmark & \checkmark & & 26.843 & 24.515 & 11.683 & 11.339 \\
\textbf{CTI (ours)} & \checkmark & \checkmark & \checkmark & \checkmark & \textbf{27.034} & \textbf{24.914} & 11.538 & \textbf{11.144} \\
\bottomrule
\end{tabular}%
}
\end{table}
\begin{figure}[t]
\centering
\setlength{\tabcolsep}{0pt}
\setlength{\lineskip}{0pt}\setlength{\lineskiplimit}{0pt}
\renewcommand{\arraystretch}{1}
\setlength{\dmw}{\dimexpr(\linewidth-3\dmg-2\dmi-\dmb)/5\relax}
\setlength{\dmbh}{\dimexpr\dmw*184/142+\dmi\relax}

\begin{tabular}{@{}c@{\hspace{\dmg}}c@{\hspace{\dmi}}c@{\hspace{\dmg}}c@{\hspace{\dmi}}c@{\hspace{\dmg}}c@{}}
& \multicolumn{2}{c}{\includegraphics[width=\dmw]{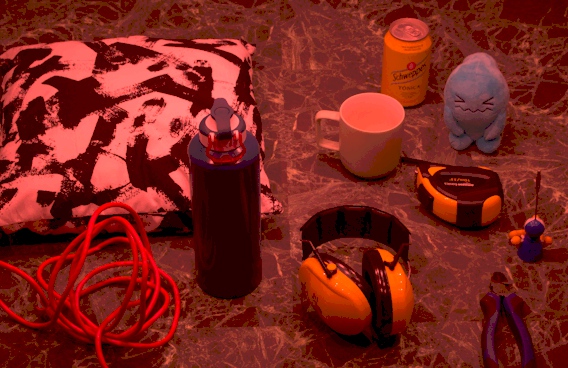}} & \multicolumn{2}{c}{\includegraphics[width=\dmw]{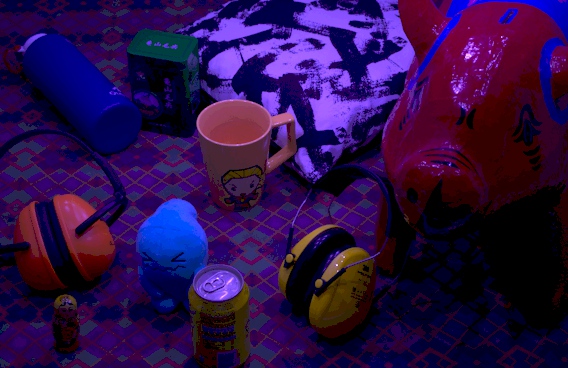}} & \\[\dmv]
& \dmhd{$\|\Delta\mathbf{z}_{ch}\|$} & \dmhd{$\|\Delta\mathbf{z}_{s}\|$}
& \dmhd{$\|\Delta\mathbf{z}_{ch}\|$} & \dmhd{$\|\Delta\mathbf{z}_{s}\|$} & \\[\dmv]
\includegraphics[width=\dmw]{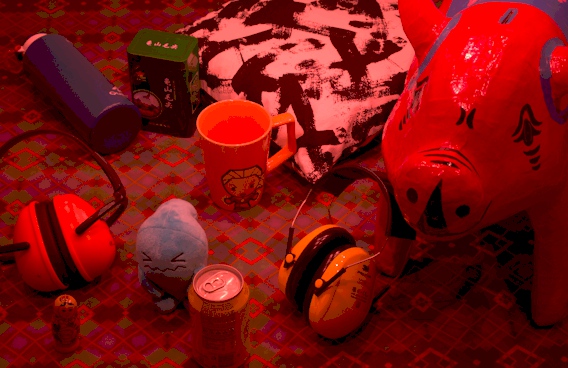} & \includegraphics[width=\dmw]{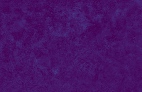} & \includegraphics[width=\dmw]{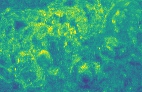}
& \includegraphics[width=\dmw]{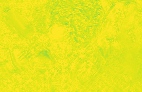} & \includegraphics[width=\dmw]{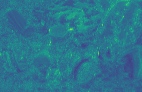} & \\[\dmi]
\includegraphics[width=\dmw]{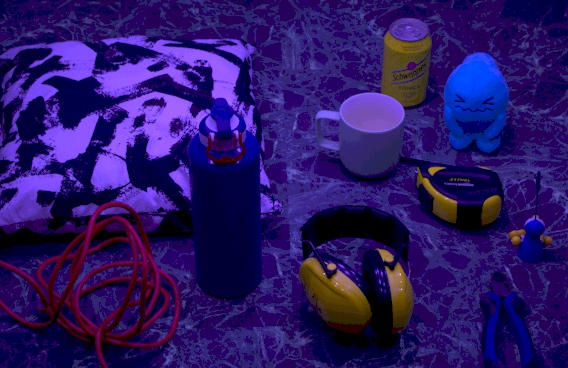} & \includegraphics[width=\dmw]{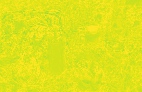} & \includegraphics[width=\dmw]{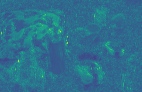}
& \includegraphics[width=\dmw]{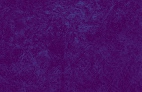} & \includegraphics[width=\dmw]{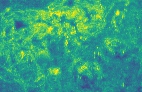} & \dmbar \\
\end{tabular}
\caption{Response of the bottleneck to a change of a single factor. Each anchor (left column) is compared against each capture (top row); every pair differs in exactly one factor, so the main diagonal changes the scene at a fixed illuminant and the anti-diagonal changes the illuminant at a fixed scene.}
\vspace{-5mm}
\label{fig:response}
\end{figure}
\label{sec:ablations}
We now isolate the contribution of each auxiliary supervision in Eq.~(6). Table~\ref{tab:ablation} reports PSNR$_L$, PSNR$_C$, MAE$_\angle$ and $\Delta$E$_{76}$ on the Nikon test set. Removing $\mathcal{L}_u$ degrades all four metrics, most notably PSNR$_C$ ($-0.40$\,dB). Further removing $\mathcal{L}_c$, which leaves TI-Retinexformer's losses (TI), costs another $0.64$\,dB PSNR$_L$, and the triplet then becomes harmful: keeping only $\mathcal{L}_i$ (I) is better on all four metrics. Since our triplet positives vary the illuminant color, $\mathcal{L}_{tr}$ pushes color out of $Z_s$, with nowhere else to go without a chromaticity channel. Removing all auxiliary terms (Reconstruction loss only) costs $2.15$\,dB PSNR$_L$ but slightly lowers MAE$_\angle$, which ignores intensity. Even so, this jointly trained model lowers $\Delta$E$_{76}$ from $14.21$ to $11.33$ relative to the TI-Retinexformer+DWB cascade: joint training brings most of the color gain, the auxiliary losses most of the fidelity gain.

\subsection{Feature Disentanglement Analysis}
\label{sec:disentanglement}
We extend the latent-space analysis of TI-Retinexformer~\cite{pilligua2025evaluating}, which separates an intensity map from a scene embedding, to the three components of our bottleneck by adding the chromaticity loss. Figure~\ref{fig:response} reports the response of $\mathbf{z}_{ch}$ and $\mathbf{z}_s$ to pairs of images that differ in exactly one factor. Each code moves with the factor it is supervised for and stays comparatively still under the other. The separation is sharpest for chromaticity. Changing the illuminant displaces $\mathbf{z}_{ch}$ more than $\mathbf{z}_{s}$ does for a change of scene, and the response is close to uniform over the frame. $\mathbf{z}_{s}$ is complementary, responding most to a change of scene, with a spatial structure that follows scene content, while still retaining a bit of response to the illuminant change.

\section{Conclusion}
\label{sec:conclusion}

We studied the interaction between low-light enhancement and auto white balance, both challenged by the weak, noisy sensor signal in low light. To evaluate them jointly, we introduced CHROMA-MILL, a RAW dataset of 50 scenes captured with a DSLR and a smartphone under four calibrated illuminants at eleven intensity levels, extended with physically valid illuminant mixtures with exact chromaticity ground truth.

Our benchmark shows that cascading low-light enhancement and AWB is insufficient. CTI-Retinexformer, which jointly supervises an illumination map, a chromaticity prediction, and a scene embedding, outperforms the best two-stage alternative by $0.7$~dB PSNR$_L$ and $1.1~\Delta E_{00}$, showing the importance of addressing both tasks jointly.

\paragraph{Limitations.}
Every CHROMA-MILL image is lit by a spatially uniform illuminant, which $\mathcal{L}_u$ builds into the model as a prior; spatially varying lighting, synthesizable with per-pixel weights in Eq.~(1), is left to future work. Our LEDs span CIE~F2 and the Planckian locus above about 3400\,K, but not warmer sources such as CIE~A. Finally, the dataset has 50 indoor scenes (8 for testing), models are trained per camera, and generalization beyond CHROMA-MILL is shown only qualitatively (Fig.~\ref{fig:external}).
\newpage
\section*{Acknowledgements} This work was supported by Grant PID2024-162555OB-I00 funded by MCIN/AEI/10.13039/501100011033 and by ERDF, EU, AIA2025-163919-C52 funded by MCIN/AEI/10.13039/501100011033, the grant Catedra ENIA UAB-Cruïlla (TSI-100929-2023-2) from the Ministry of Economic Affairs and Digital Transformation of Spain, and by the Generalitat de Catalunya CERCA Program. JVC also acknowledges the 2025 Leonardo Grant for Scientific Research and Cultural Creation from the BBVA Foundation. The BBVA Foundation accepts no responsibility for the opinions, statements and contents included in the project and/or the results thereof, which are entirely the responsibility of the authors. DSL also acknowledges the FPI grant from Spanish Ministry of Science and Innovation (PRE2022-101525).

{
    \small
    \bibliographystyle{ieeenat_fullname}
    \bibliography{chromamill}
}

\appendix
\twocolumn[\centering\Large\bfseries Supplementary Material\par\vspace{1.5em}]
\begin{figure*}[t!]
    \centering
    \includegraphics[width=\linewidth]{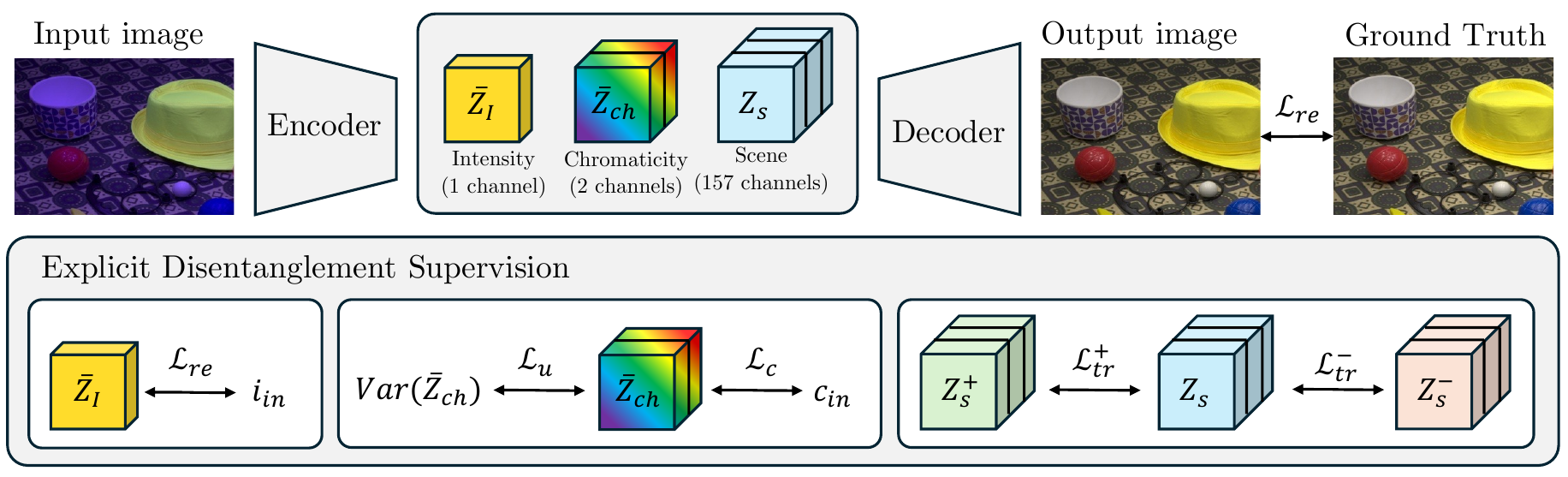}
    \caption{\textbf{Overview of CTI-Retinexformer.}}    \label{fig:method}
\end{figure*}

We provide additional material to supplement our main submission, covering:

\begin{enumerate}
    \item Extended method details.
    \item Additional quantitative results.
\end{enumerate}

\section{Extended Method Details}
Figure~\ref{fig:method} depicts an overview of CTI-Retinexformer. Given a low-light input image $x$, the encoder maps the image into a latent representation that is explicitly disentangled into three components: an intensity representation $Z_I$, a chromaticity representation $Z_{ch}$, and a scene-content representation $Z_s$.
Following TI-Retinexformer~\cite{pilligua2025evaluating}, $Z_I$ consists of a single channel and encodes the normalized illumination intensity of the input. We extend this decomposition with a two-channel chromaticity representation $Z_{ch}$, which predicts the CIE $xy$ chromaticity of the scene illuminant, while the remaining $157$ channels form $Z_s$ and encode scene content independently of illumination.
The three representations are jointly processed by the decoder to reconstruct the enhanced and white-balanced output $\hat{y}$.
Each latent component is explicitly supervised during training. The intensity loss $\mathcal{L}_{i}$ encourages the spatial mean of $Z_I$ to match the normalized light-power level of the input, $i_{in}=B/254$. The chromaticity loss $\mathcal{L}_{c}$ supervises $Z_{ch}$ using the ground-truth illuminant chromaticity, while the uniformity loss $\mathcal{L}_{u}$ minimizes its spatial variance to enforce the spatially uniform illuminant prior. Finally, the scene-content loss $\mathcal{L}_{tr}$ uses a triplet objective to encourage $Z_s$ to remain invariant to changes in illumination intensity and color while preserving scene identity.
The reconstruction loss $\mathcal{L}_{re}$ supervises the final output as a mean L1 distance with the same weight for every input level.
The complete training objective is the sum of the reconstruction and four auxiliary losses:
$\mathcal{L} = \mathcal{L}_{re} + \mathcal{L}_{i} + \mathcal{L}_{c} + \mathcal{L}_{u} + \mathcal{L}_{tr}$.

\section{Additional Quantitative Results}
\label{sec:supp}
Tables~\ref{tab:nikon_results} and~\ref{tab:phone_results} report the complete evaluation, with both cascade orderings listed separately for every two-stage pipeline for the DSLR and the smartphone camera, respectively.
Table~\ref{tab:levels_supp} then breaks the DSLR evaluation down by input intensity level, reporting every method at each of the ten levels.

\begin{table*}[t]
\centering
\caption{Complete evaluation on the Augmented (blended-illuminant) Nikon test set, with both cascade orderings of every two-stage pipeline. White-balance-first applies the white-balance module to the raw low-light input; enhancement-first applies it to the enhancer's output. The main paper reports the better of the two per pipeline. Best results are highlighted in \colorbox{cbest}{red}, second in \colorbox{csecond}{orange}, and third in \colorbox{cthird}{yellow}.}
\label{tab:nikon_results}
\resizebox{\textwidth}{!}{%
\begin{tabular}{l l ccccccc}
\toprule
 & Order & PSNR$_L$ $\uparrow$ & PSNR$_C$ $\uparrow$ & SSIM $\uparrow$ & LPIPS $\downarrow$ & MAE$_{\angle}$ $\downarrow$ & $\Delta$E$_{76}$ $\downarrow$ & $\Delta$E$_{00}$ $\downarrow$ \\
\midrule
Unprocessed & -- & 18.764 & 17.676 & 0.552 & 0.253 & 23.160 & 22.843 & 16.152 \\
\addlinespace[2pt]
\multirow{2}{*}{TI-Retinexformer + DWB} & WB-first & 24.776 & 22.677 & 0.759 & 0.172 & 14.260 & 14.206 & 10.178 \\
 & LLIE-first & 21.092 & 18.821 & 0.643 & 0.289 & 18.933 & 20.162 & 13.871 \\
\addlinespace[2pt]
\multirow{2}{*}{TI-Retinexformer + FC4} & WB-first & 23.784 & 19.546 & 0.653 & 0.280 & 26.035 & 24.476 & 16.652 \\
 & LLIE-first & 20.264 & 18.958 & 0.602 & 0.275 & 26.055 & 22.999 & 16.907 \\
\addlinespace[2pt]
\multirow{2}{*}{TI-Retinexformer + FFCC} & WB-first & 24.235 & 19.280 & 0.660 & 0.262 & 27.544 & 25.939 & 17.328 \\
 & LLIE-first & 23.723 & 19.638 & 0.677 & 0.253 & 26.506 & 24.429 & 16.861 \\
\addlinespace[2pt]
\multirow{2}{*}{TI-Retinexformer + RL-AWB} & WB-first & 24.972 & 22.129 & 0.696 & 0.200 & 19.126 & 18.723 & 12.568 \\
 & LLIE-first & 22.915 & 22.034 & 0.685 & 0.183 & 13.618 & 14.603 & 10.792 \\
\addlinespace[2pt]
\multirow{2}{*}{Retinexformer + DWB} & WB-first & 23.226 & 20.059 & 0.626 & 0.286 & 21.015 & 20.133 & 13.381 \\
 & LLIE-first & 22.525 & 19.757 & 0.671 & 0.274 & 18.756 & 18.894 & 13.079 \\
\addlinespace[2pt]
\multirow{2}{*}{Retinexformer + FC4} & WB-first & 23.599 & 19.875 & 0.649 & 0.277 & 26.750 & 23.853 & 16.555 \\
 & LLIE-first & 20.036 & 18.873 & 0.603 & 0.273 & 26.109 & 22.481 & 16.901 \\
\addlinespace[2pt]
\multirow{2}{*}{Retinexformer + FFCC} & WB-first & 24.566 & 19.700 & 0.662 & 0.257 & 27.464 & 24.672 & 16.751 \\
 & LLIE-first & 23.397 & 19.818 & 0.673 & 0.252 & 27.539 & 24.137 & 16.922 \\
\addlinespace[2pt]
\multirow{2}{*}{Retinexformer + RL-AWB} & WB-first & 25.666 & 22.885 & 0.702 & 0.195 & 19.149 & 17.665 & 12.041 \\
 & LLIE-first & 22.962 & 22.085 & 0.688 & 0.173 & \cellcolor{cthird}12.239 & 13.824 & 10.259 \\
\addlinespace[2pt]
\multirow{2}{*}{DarkIR + DWB} & WB-first & \cellcolor{cthird}26.087 & \cellcolor{csecond}23.676 & \cellcolor{cthird}0.783 & 0.165 & 14.738 & 13.787 & 9.838 \\
 & LLIE-first & 22.740 & 19.804 & 0.677 & 0.289 & 18.852 & 18.694 & 13.110 \\
\addlinespace[2pt]
\multirow{2}{*}{DarkIR + FC4} & WB-first & 23.787 & 20.102 & 0.664 & 0.283 & 26.685 & 23.934 & 16.646 \\
 & LLIE-first & 20.548 & 19.239 & 0.622 & 0.272 & 25.926 & 22.302 & 16.755 \\
\addlinespace[2pt]
\multirow{2}{*}{DarkIR + FFCC} & WB-first & 24.882 & 20.019 & 0.680 & 0.260 & 27.671 & 24.858 & 16.763 \\
 & LLIE-first & 23.985 & 20.160 & 0.694 & 0.256 & 27.660 & 24.912 & 17.292 \\
\addlinespace[2pt]
\multirow{2}{*}{DarkIR + RL-AWB} & WB-first & 25.578 & 22.515 & 0.718 & 0.198 & 19.399 & 18.224 & 12.282 \\
 & LLIE-first & 23.486 & 22.470 & 0.706 & 0.178 & 13.259 & 14.133 & 10.629 \\
\addlinespace[2pt]
\multirow{2}{*}{MIRNet + DWB} & WB-first & \cellcolor{csecond}26.350 & \cellcolor{cthird}23.663 & \cellcolor{csecond}0.787 & 0.165 & 14.847 & \cellcolor{cthird}13.510 & \cellcolor{csecond}9.697 \\
 & LLIE-first & 25.295 & 22.704 & 0.739 & 0.200 & 18.058 & 15.698 & 11.545 \\
\addlinespace[2pt]
\multirow{2}{*}{MIRNet + FC4} & WB-first & 23.716 & 19.959 & 0.660 & 0.273 & 26.720 & 23.663 & 16.385 \\
 & LLIE-first & 20.543 & 18.996 & 0.616 & 0.261 & 26.177 & 22.016 & 16.541 \\
\addlinespace[2pt]
\multirow{2}{*}{MIRNet + FFCC} & WB-first & 24.525 & 19.759 & 0.673 & 0.253 & 27.878 & 24.744 & 16.755 \\
 & LLIE-first & 23.426 & 19.821 & 0.685 & 0.245 & 27.689 & 24.129 & 16.851 \\
\addlinespace[2pt]
\multirow{2}{*}{MIRNet + RL-AWB} & WB-first & 25.429 & 22.604 & 0.712 & 0.192 & 19.636 & 17.935 & 12.175 \\
 & LLIE-first & 22.761 & 21.861 & 0.687 & 0.168 & 12.924 & 14.040 & 10.473 \\
\addlinespace[2pt]
\multirow{2}{*}{PromptNorm + DWB} & WB-first & 21.922 & 20.625 & 0.774 & 0.164 & 13.934 & 16.024 & 11.674 \\
 & LLIE-first & 22.235 & 20.855 & 0.773 & \cellcolor{cthird}0.157 & 13.495 & 15.173 & 11.221 \\
\addlinespace[2pt]
\multirow{2}{*}{PromptNorm + FC4} & WB-first & 24.828 & 19.087 & 0.698 & 0.274 & 26.244 & 26.641 & 17.875 \\
 & LLIE-first & 22.886 & 19.909 & 0.678 & 0.251 & 25.027 & 21.959 & 16.531 \\
\addlinespace[2pt]
\multirow{2}{*}{PromptNorm + FFCC} & WB-first & 25.049 & 18.972 & 0.713 & 0.257 & 27.389 & 27.734 & 18.119 \\
 & LLIE-first & 25.578 & 19.813 & 0.732 & 0.240 & 26.498 & 25.973 & 17.606 \\
\addlinespace[2pt]
\multirow{2}{*}{PromptNorm + RL-AWB} & WB-first & 23.435 & 20.469 & 0.736 & 0.191 & 18.688 & 20.577 & 13.724 \\
 & LLIE-first & 24.864 & 23.429 & 0.753 & \cellcolor{csecond}0.153 & \cellcolor{csecond}11.812 & \cellcolor{csecond}12.618 & \cellcolor{cthird}9.714 \\
\addlinespace[2pt]
\cdashline{1-9}
Ours (CTI-Retinexformer) & -- & \cellcolor{cbest}27.034 & \cellcolor{cbest}24.914 & \cellcolor{cbest}0.809 & \cellcolor{cbest}0.134 & \cellcolor{cbest}11.538 & \cellcolor{cbest}11.144 & \cellcolor{cbest}8.562 \\
\bottomrule
\end{tabular}%
}
\end{table*}
\begin{table*}[t]
\centering
\caption{Complete evaluation on the Augmented (blended-illuminant) phone test set, with both cascade orderings of every two-stage pipeline. White-balance-first applies the white-balance module to the raw low-light input; enhancement-first applies it to the enhancer's output. The main paper reports the better of the two per pipeline. Best results are highlighted in \colorbox{cbest}{red}, second in \colorbox{csecond}{orange}, and third in \colorbox{cthird}{yellow}.}
\label{tab:phone_results}
\resizebox{\textwidth}{!}{%
\begin{tabular}{l l ccccccc}
\toprule
 & Order & PSNR$_L$ $\uparrow$ & PSNR$_C$ $\uparrow$ & SSIM $\uparrow$ & LPIPS $\downarrow$ & MAE$_{\angle}$ $\downarrow$ & $\Delta$E$_{76}$ $\downarrow$ & $\Delta$E$_{00}$ $\downarrow$ \\
\midrule
Unprocessed & -- & 16.634 & 15.761 & 0.439 & 0.284 & 23.062 & 27.275 & 19.841 \\
\addlinespace[2pt]
\multirow{2}{*}{Retinexformer + DWB} & WB-first & \cellcolor{cthird}24.090 & \cellcolor{cthird}21.931 & 0.626 & 0.178 & 12.295 & \cellcolor{cthird}13.301 & \cellcolor{cthird}9.804 \\
 & LLIE-first & \cellcolor{cbest}24.421 & \cellcolor{csecond}22.170 & \cellcolor{csecond}0.647 & \cellcolor{csecond}0.166 & \cellcolor{csecond}11.440 & \cellcolor{csecond}12.463 & \cellcolor{csecond}9.327 \\
\addlinespace[2pt]
\multirow{2}{*}{Retinexformer + FC4} & WB-first & 23.662 & 20.602 & 0.601 & 0.232 & 18.286 & 18.444 & 14.303 \\
 & LLIE-first & 19.845 & 18.500 & 0.561 & 0.232 & 19.372 & 20.579 & 16.181 \\
\addlinespace[2pt]
\multirow{2}{*}{Retinexformer + FFCC} & WB-first & 22.048 & 19.512 & 0.584 & 0.243 & 22.342 & 25.213 & 17.334 \\
 & LLIE-first & 20.552 & 18.845 & 0.584 & 0.244 & 22.978 & 25.388 & 18.214 \\
\addlinespace[2pt]
\multirow{2}{*}{Retinexformer + RL-AWB} & WB-first & 23.457 & 21.774 & 0.611 & 0.184 & 12.924 & 14.797 & 11.068 \\
 & LLIE-first & 19.046 & 18.626 & 0.577 & 0.185 & \cellcolor{cthird}12.027 & 18.182 & 13.798 \\
\addlinespace[2pt]
\multirow{2}{*}{DarkIR + DWB} & WB-first & 23.235 & 21.304 & 0.613 & 0.172 & 12.908 & 14.048 & 10.247 \\
 & LLIE-first & 23.597 & 21.545 & 0.627 & \cellcolor{cthird}0.167 & 12.249 & 13.395 & 9.970 \\
\addlinespace[2pt]
\multirow{2}{*}{DarkIR + FC4} & WB-first & 23.470 & 19.943 & 0.578 & 0.237 & 20.197 & 20.572 & 15.449 \\
 & LLIE-first & 19.275 & 17.895 & 0.525 & 0.236 & 20.314 & 21.670 & 16.649 \\
\addlinespace[2pt]
\multirow{2}{*}{DarkIR + FFCC} & WB-first & 21.979 & 19.091 & 0.564 & 0.244 & 23.499 & 26.478 & 17.752 \\
 & LLIE-first & 20.391 & 18.613 & 0.562 & 0.243 & 24.207 & 26.351 & 18.418 \\
\addlinespace[2pt]
\multirow{2}{*}{DarkIR + RL-AWB} & WB-first & 23.652 & 21.267 & 0.593 & 0.186 & 14.558 & 16.426 & 11.776 \\
 & LLIE-first & 18.859 & 18.407 & 0.551 & 0.183 & 13.072 & 18.509 & 13.985 \\
\addlinespace[2pt]
\multirow{2}{*}{MIRNet + DWB} & WB-first & 23.247 & 21.276 & 0.627 & 0.190 & 12.658 & 14.035 & 10.305 \\
 & LLIE-first & 23.330 & 21.362 & \cellcolor{cthird}0.632 & 0.181 & 12.103 & 13.626 & 10.092 \\
\addlinespace[2pt]
\multirow{2}{*}{MIRNet + FC4} & WB-first & 21.679 & 19.336 & 0.563 & 0.250 & 20.912 & 21.083 & 16.048 \\
 & LLIE-first & 18.439 & 17.380 & 0.517 & 0.249 & 20.222 & 22.329 & 17.111 \\
\addlinespace[2pt]
\multirow{2}{*}{MIRNet + FFCC} & WB-first & 20.781 & 18.659 & 0.555 & 0.251 & 23.952 & 26.775 & 18.243 \\
 & LLIE-first & 19.169 & 17.967 & 0.550 & 0.257 & 24.477 & 26.726 & 19.035 \\
\addlinespace[2pt]
\multirow{2}{*}{MIRNet + RL-AWB} & WB-first & 22.709 & 21.017 & 0.591 & 0.199 & 15.218 & 17.125 & 12.425 \\
 & LLIE-first & 17.981 & 17.702 & 0.535 & 0.203 & 13.285 & 19.774 & 14.902 \\
\addlinespace[2pt]
\multirow{2}{*}{PromptNorm + DWB} & WB-first & 21.187 & 19.713 & 0.613 & 0.185 & 12.357 & 15.827 & 11.715 \\
 & LLIE-first & 21.834 & 20.121 & 0.622 & 0.181 & 12.256 & 15.146 & 11.344 \\
\addlinespace[2pt]
\multirow{2}{*}{PromptNorm + FC4} & WB-first & 22.504 & 19.196 & 0.582 & 0.247 & 19.363 & 21.610 & 16.139 \\
 & LLIE-first & 20.157 & 18.194 & 0.547 & 0.247 & 20.177 & 21.585 & 16.584 \\
\addlinespace[2pt]
\multirow{2}{*}{PromptNorm + FFCC} & WB-first & 21.649 & 18.523 & 0.571 & 0.256 & 22.909 & 27.766 & 18.443 \\
 & LLIE-first & 21.340 & 18.886 & 0.577 & 0.257 & 23.691 & 26.671 & 18.516 \\
\addlinespace[2pt]
\multirow{2}{*}{PromptNorm + RL-AWB} & WB-first & 22.250 & 20.238 & 0.592 & 0.199 & 13.978 & 17.439 & 12.742 \\
 & LLIE-first & 19.614 & 18.753 & 0.569 & 0.198 & 13.239 & 18.142 & 13.770 \\
\addlinespace[2pt]
\cdashline{1-9}
Ours (CTI-Retinexformer) & -- & \cellcolor{csecond}24.352 & \cellcolor{cbest}22.776 & \cellcolor{cbest}0.649 & \cellcolor{cbest}0.136 & \cellcolor{cbest}10.505 & \cellcolor{cbest}11.387 & \cellcolor{cbest}8.881 \\
\bottomrule
\end{tabular}%
}
\end{table*}
\clearpage
\onecolumn
\footnotesize
\setlength{\tabcolsep}{3pt}
\begin{longtable}{l cccc cccc cccc}
\caption{Performance across all ten input intensity levels on the DSLR test set. Level~1 is the darkest input and level~10 the brightest; the reference is captured at maximum brightness. Every method of the main comparison is listed at its selected cascade ordering, rows marked $^{\dagger}$ enhancement-first and the rest white-balance-first. Within each level, \colorbox{cbest}{best}, \colorbox{csecond}{second best} and \colorbox{cthird}{third best} results are highlighted per metric. Level-averaged results are in the main paper.}
\label{tab:levels_supp}\\
\endfirsthead
\multicolumn{13}{c}{\tablename~\thetable{} -- continued from the previous page}\\
\endhead
\toprule
 & \multicolumn{4}{c}{Level 1} & \multicolumn{4}{c}{Level 2} & \multicolumn{4}{c}{Level 3} \\
\cmidrule(lr){2-5}\cmidrule(lr){6-9}\cmidrule(lr){10-13}
DSLR & PSNR$_L$ $\uparrow$ & PSNR$_C$ $\uparrow$ & MAE$_{\angle}$ $\downarrow$ & $\Delta$E$_{00}$ $\downarrow$ & PSNR$_L$ $\uparrow$ & PSNR$_C$ $\uparrow$ & MAE$_{\angle}$ $\downarrow$ & $\Delta$E$_{00}$ $\downarrow$ & PSNR$_L$ $\uparrow$ & PSNR$_C$ $\uparrow$ & MAE$_{\angle}$ $\downarrow$ & $\Delta$E$_{00}$ $\downarrow$ \\
\midrule
Unprocessed & 11.984 & 11.842 & \cellcolor{cbest}23.318 & 21.523 & 13.524 & 13.435 & 25.303 & 19.416 & 14.863 & 14.740 & 25.497 & 17.980 \\
\hdashline
\addlinespace[2pt]
TI-Retinexformer + DWB & 18.493 & 17.223 & 25.359 & \cellcolor{cthird}16.058 & 23.470 & 20.873 & 20.340 & 12.876 & 24.990 & 22.543 & 16.114 & 10.694 \\
TI-Retinexformer + FC4 & 16.048 & 14.839 & 28.792 & 18.541 & 20.430 & 17.811 & 29.190 & 17.604 & 22.616 & 19.043 & 27.831 & 17.018 \\
TI-Retinexformer + FFCC$^{\dagger}$ & 17.858 & 16.776 & 23.990 & 16.571 & 21.007 & 18.381 & 28.715 & 17.538 & 22.852 & 19.338 & 28.432 & 17.258 \\
TI-Retinexformer + RL-AWB$^{\dagger}$ & 16.761 & 16.245 & 24.437 & 17.079 & 20.097 & 19.451 & 18.692 & 13.630 & 21.557 & 20.830 & 16.686 & 12.199 \\
\addlinespace[2pt]
Retinexformer + DWB$^{\dagger}$ & 19.591 & 17.108 & 26.730 & 16.392 & 22.274 & 19.464 & 22.248 & 13.828 & 22.475 & 19.718 & 20.063 & 13.230 \\
Retinexformer + FC4 & 15.853 & 14.657 & 32.633 & 18.989 & 20.477 & 17.966 & 30.889 & 17.763 & 22.476 & 19.190 & 28.615 & 17.087 \\
Retinexformer + FFCC$^{\dagger}$ & 18.122 & 16.817 & 31.245 & 18.362 & 20.354 & 18.324 & 30.325 & 17.874 & 21.995 & 19.233 & 28.831 & 17.336 \\
Retinexformer + RL-AWB & 16.088 & 15.013 & 28.937 & 17.756 & 21.347 & 18.673 & 28.445 & 16.177 & 23.546 & 20.134 & 25.933 & 15.330 \\
\addlinespace[2pt]
DarkIR + DWB & 19.880 & 18.152 & 30.087 & 18.462 & 23.870 & 21.058 & 20.968 & 13.236 & \cellcolor{cthird}26.341 & \cellcolor{csecond}23.486 & 16.057 & \cellcolor{csecond}10.153 \\
DarkIR + FC4$^{\dagger}$ & 16.179 & 15.912 & 37.630 & 22.442 & 18.884 & 18.258 & 27.533 & 17.502 & 19.561 & 18.767 & 25.074 & 16.437 \\
DarkIR + FFCC$^{\dagger}$ & 18.551 & 17.452 & 32.709 & 22.091 & 21.554 & 18.575 & 31.198 & 18.860 & 22.405 & 19.571 & 29.061 & 17.349 \\
DarkIR + RL-AWB$^{\dagger}$ & 17.664 & 17.233 & 27.645 & 19.048 & 20.625 & 19.952 & 19.302 & 13.567 & 21.683 & 20.957 & 15.027 & 11.534 \\
\addlinespace[2pt]
MIRNet + DWB & 20.938 & 18.176 & 30.293 & 17.160 & \cellcolor{csecond}24.741 & \cellcolor{csecond}21.377 & 21.340 & \cellcolor{cthird}12.527 & \cellcolor{csecond}26.883 & \cellcolor{cthird}23.413 & 16.451 & \cellcolor{cthird}10.188 \\
MIRNet + FC4 & 16.566 & 15.158 & 33.910 & 18.918 & 19.825 & 17.702 & 31.553 & 17.713 & 22.137 & 19.145 & 28.617 & 16.800 \\
MIRNet + FFCC$^{\dagger}$ & 18.242 & 16.859 & 33.236 & 18.396 & 20.191 & 18.132 & 31.128 & 17.694 & 21.830 & 19.191 & 28.830 & 17.061 \\
MIRNet + RL-AWB & 17.393 & 15.779 & 32.955 & 17.616 & 21.159 & 18.421 & 29.618 & 16.436 & 23.401 & 19.952 & 26.277 & 15.449 \\
\addlinespace[2pt]
PromptNorm + DWB$^{\dagger}$ & \cellcolor{csecond}22.428 & \cellcolor{csecond}19.834 & \cellcolor{csecond}23.328 & \cellcolor{cbest}15.707 & 21.451 & 20.052 & \cellcolor{csecond}16.532 & 12.777 & 21.901 & 20.560 & \cellcolor{cthird}14.358 & 11.645 \\
PromptNorm + FC4$^{\dagger}$ & 18.391 & 17.028 & 33.871 & 21.282 & 20.933 & 18.871 & 27.857 & 18.272 & 22.053 & 19.555 & 25.464 & 16.931 \\
PromptNorm + FFCC$^{\dagger}$ & \cellcolor{cthird}22.173 & \cellcolor{cthird}18.749 & 30.536 & 20.008 & \cellcolor{cthird}24.287 & 19.294 & 27.975 & 18.457 & 25.137 & 19.585 & 27.076 & 17.838 \\
PromptNorm + RL-AWB$^{\dagger}$ & 20.046 & 18.696 & \cellcolor{cthird}23.850 & 16.751 & 22.562 & \cellcolor{cthird}21.230 & \cellcolor{cthird}16.714 & \cellcolor{csecond}12.259 & 23.666 & 22.367 & \cellcolor{csecond}13.578 & 10.649 \\
\hdashline
\addlinespace[2pt]
Ours (CTI-Retinexformer) & \cellcolor{cbest}22.432 & \cellcolor{cbest}20.117 & 25.771 & \cellcolor{csecond}16.023 & \cellcolor{cbest}25.617 & \cellcolor{cbest}23.457 & \cellcolor{cbest}15.781 & \cellcolor{cbest}10.588 & \cellcolor{cbest}27.389 & \cellcolor{cbest}25.049 & \cellcolor{cbest}12.425 & \cellcolor{cbest}8.737 \\
\bottomrule
\addlinespace[10pt]
\toprule
 & \multicolumn{4}{c}{Level 4} & \multicolumn{4}{c}{Level 5} & \multicolumn{4}{c}{Level 6} \\
\cmidrule(lr){2-5}\cmidrule(lr){6-9}\cmidrule(lr){10-13}
DSLR & PSNR$_L$ $\uparrow$ & PSNR$_C$ $\uparrow$ & MAE$_{\angle}$ $\downarrow$ & $\Delta$E$_{00}$ $\downarrow$ & PSNR$_L$ $\uparrow$ & PSNR$_C$ $\uparrow$ & MAE$_{\angle}$ $\downarrow$ & $\Delta$E$_{00}$ $\downarrow$ & PSNR$_L$ $\uparrow$ & PSNR$_C$ $\uparrow$ & MAE$_{\angle}$ $\downarrow$ & $\Delta$E$_{00}$ $\downarrow$ \\
\midrule
Unprocessed & 16.259 & 16.042 & 24.507 & 16.801 & 17.726 & 17.317 & 23.478 & 15.750 & 19.107 & 18.435 & 22.842 & 15.031 \\
\hdashline
\addlinespace[2pt]
TI-Retinexformer + DWB & 25.119 & 22.921 & 13.669 & 9.849 & 25.520 & 23.384 & 12.411 & 9.284 & 25.867 & 23.746 & 11.622 & 8.870 \\
TI-Retinexformer + FC4 & 23.895 & 19.811 & 26.555 & 16.582 & 24.708 & 20.184 & 25.743 & 16.389 & 25.235 & 20.478 & 25.116 & 16.231 \\
TI-Retinexformer + FFCC$^{\dagger}$ & 23.512 & 19.650 & 27.597 & 17.141 & 24.170 & 19.930 & 26.956 & 16.929 & 24.640 & 20.126 & 26.537 & 16.822 \\
TI-Retinexformer + RL-AWB$^{\dagger}$ & 22.615 & 21.736 & 14.622 & 11.146 & 23.404 & 22.477 & 12.678 & 10.227 & 23.924 & 22.951 & 11.552 & 9.696 \\
\addlinespace[2pt]
Retinexformer + DWB$^{\dagger}$ & 22.766 & 19.966 & 18.626 & 12.825 & 22.906 & 20.167 & 17.720 & 12.587 & 22.924 & 20.190 & 17.204 & 12.512 \\
Retinexformer + FC4 & 23.620 & 20.030 & 26.982 & 16.516 & 24.430 & 20.544 & 25.973 & 16.211 & 25.021 & 20.880 & 25.197 & 16.011 \\
Retinexformer + FFCC$^{\dagger}$ & 23.143 & 19.850 & 27.608 & 16.910 & 23.852 & 20.168 & 27.055 & 16.669 & 24.179 & 20.334 & 26.743 & 16.600 \\
Retinexformer + RL-AWB & 25.091 & 21.645 & 23.039 & 14.000 & 26.420 & 22.960 & 20.335 & 12.669 & 27.180 & 23.805 & 18.459 & 11.762 \\
\addlinespace[2pt]
DarkIR + DWB & \cellcolor{csecond}27.072 & \cellcolor{csecond}24.274 & 13.689 & \cellcolor{csecond}9.018 & \cellcolor{csecond}27.337 & \cellcolor{csecond}24.743 & 12.349 & \cellcolor{csecond}8.421 & \cellcolor{cthird}27.407 & \cellcolor{csecond}24.975 & 11.502 & \cellcolor{csecond}8.048 \\
DarkIR + FC4$^{\dagger}$ & 20.386 & 19.361 & 24.111 & 15.906 & 21.109 & 19.733 & 23.951 & 15.762 & 21.280 & 19.782 & 24.048 & 15.845 \\
DarkIR + FFCC$^{\dagger}$ & 23.782 & 20.257 & 27.724 & 16.829 & 24.738 & 20.605 & 26.914 & 16.581 & 24.930 & 20.764 & 26.436 & 16.419 \\
DarkIR + RL-AWB$^{\dagger}$ & 23.068 & 22.070 & \cellcolor{cthird}13.006 & 10.522 & 24.055 & 22.988 & \cellcolor{cthird}11.323 & 9.592 & 24.570 & 23.428 & \cellcolor{cthird}10.371 & 9.119 \\
\addlinespace[2pt]
MIRNet + DWB & \cellcolor{cthird}26.969 & \cellcolor{cthird}24.163 & 13.779 & \cellcolor{cthird}9.141 & \cellcolor{cthird}26.990 & \cellcolor{cthird}24.442 & 12.306 & \cellcolor{cthird}8.639 & \cellcolor{csecond}27.415 & \cellcolor{cthird}24.906 & 11.486 & \cellcolor{cthird}8.117 \\
MIRNet + FC4 & 23.639 & 20.170 & 26.620 & 16.170 & 24.639 & 20.695 & 25.518 & 15.920 & 25.297 & 21.051 & 24.809 & 15.778 \\
MIRNet + FFCC$^{\dagger}$ & 23.261 & 19.909 & 27.539 & 16.814 & 23.814 & 20.192 & 26.989 & 16.674 & 24.282 & 20.448 & 26.515 & 16.531 \\
MIRNet + RL-AWB & 24.915 & 21.470 & 23.135 & 14.133 & 26.312 & 22.775 & 20.318 & 12.730 & 26.816 & 23.513 & 18.323 & 11.859 \\
\addlinespace[2pt]
PromptNorm + DWB$^{\dagger}$ & 22.275 & 20.947 & 13.086 & 10.940 & 22.402 & 21.139 & 12.208 & 10.514 & 22.458 & 21.235 & 11.628 & 10.260 \\
PromptNorm + FC4$^{\dagger}$ & 22.957 & 20.013 & 24.160 & 16.191 & 23.481 & 20.279 & 23.535 & 15.746 & 23.737 & 20.391 & 23.250 & 15.555 \\
PromptNorm + FFCC$^{\dagger}$ & 25.691 & 19.842 & 26.425 & 17.486 & 25.935 & 19.965 & 26.078 & 17.276 & 26.066 & 20.025 & 25.831 & 17.161 \\
PromptNorm + RL-AWB$^{\dagger}$ & 24.718 & 23.324 & \cellcolor{csecond}11.572 & 9.545 & 25.365 & 23.920 & \cellcolor{csecond}10.213 & 8.872 & 25.737 & 24.287 & \cellcolor{csecond}9.291 & 8.381 \\
\hdashline
\addlinespace[2pt]
Ours (CTI-Retinexformer) & \cellcolor{cbest}28.057 & \cellcolor{cbest}25.667 & \cellcolor{cbest}10.607 & \cellcolor{cbest}7.870 & \cellcolor{cbest}28.216 & \cellcolor{cbest}25.924 & \cellcolor{cbest}9.476 & \cellcolor{cbest}7.362 & \cellcolor{cbest}28.156 & \cellcolor{cbest}26.003 & \cellcolor{cbest}8.778 & \cellcolor{cbest}7.093 \\
\bottomrule
\newpage
\addlinespace[10pt]
\toprule
 & \multicolumn{4}{c}{Level 7} & \multicolumn{4}{c}{Level 8} & \multicolumn{4}{c}{Level 9} \\
\cmidrule(lr){2-5}\cmidrule(lr){6-9}\cmidrule(lr){10-13}
DSLR & PSNR$_L$ $\uparrow$ & PSNR$_C$ $\uparrow$ & MAE$_{\angle}$ $\downarrow$ & $\Delta$E$_{00}$ $\downarrow$ & PSNR$_L$ $\uparrow$ & PSNR$_C$ $\uparrow$ & MAE$_{\angle}$ $\downarrow$ & $\Delta$E$_{00}$ $\downarrow$ & PSNR$_L$ $\uparrow$ & PSNR$_C$ $\uparrow$ & MAE$_{\angle}$ $\downarrow$ & $\Delta$E$_{00}$ $\downarrow$ \\
\midrule
Unprocessed & 20.650 & 19.565 & 22.388 & 14.436 & 22.837 & 21.019 & 21.602 & 13.716 & 24.396 & 21.777 & 21.688 & 13.652 \\
\hdashline
\addlinespace[2pt]
TI-Retinexformer + DWB & 26.144 & 24.023 & 11.095 & 8.587 & 26.219 & 24.091 & 10.800 & 8.513 & 25.883 & 23.876 & 10.555 & 8.567 \\
TI-Retinexformer + FC4 & 25.663 & 20.664 & 24.722 & 16.142 & 26.251 & 20.898 & 24.055 & 15.915 & 26.457 & 20.729 & 24.457 & 16.268 \\
TI-Retinexformer + FFCC$^{\dagger}$ & 25.073 & 20.279 & 26.231 & 16.738 & 25.880 & 20.618 & 25.564 & 16.477 & 25.972 & 20.465 & 25.919 & 16.835 \\
TI-Retinexformer + RL-AWB$^{\dagger}$ & 24.421 & 23.453 & 10.145 & 8.949 & 25.215 & 24.197 & 9.305 & 8.445 & 25.387 & 24.336 & 9.131 & 8.425 \\
\addlinespace[2pt]
Retinexformer + DWB$^{\dagger}$ & 23.055 & 20.226 & 16.821 & 12.454 & 23.140 & 20.295 & 16.164 & 12.282 & 23.182 & 20.285 & 15.879 & 12.251 \\
Retinexformer + FC4 & 25.470 & 21.116 & 24.778 & 15.894 & 26.040 & 21.446 & 24.040 & 15.600 & 26.257 & 21.350 & 24.447 & 15.936 \\
Retinexformer + FFCC$^{\dagger}$ & 24.706 & 20.582 & 26.354 & 16.482 & 25.530 & 20.901 & 25.761 & 16.265 & 25.886 & 20.838 & 26.055 & 16.587 \\
Retinexformer + RL-AWB & \cellcolor{cbest}28.103 & \cellcolor{csecond}25.280 & 15.106 & 9.983 & \cellcolor{cbest}29.455 & \cellcolor{cbest}26.962 & 11.024 & 7.807 & \cellcolor{cbest}29.619 & \cellcolor{cbest}27.128 & 10.068 & \cellcolor{csecond}7.466 \\
\addlinespace[2pt]
DarkIR + DWB & 27.431 & 25.092 & 10.982 & \cellcolor{csecond}7.809 & 27.391 & 25.114 & 10.700 & \cellcolor{cthird}7.718 & 27.197 & 24.986 & 10.465 & 7.721 \\
DarkIR + FC4$^{\dagger}$ & 21.483 & 19.871 & 24.177 & 15.941 & 22.035 & 20.260 & 23.943 & 15.737 & 22.073 & 20.086 & 24.603 & 16.146 \\
DarkIR + FFCC$^{\dagger}$ & 25.092 & 20.880 & 26.107 & 16.324 & 25.941 & 21.184 & 25.440 & 16.048 & 26.130 & 21.007 & 25.865 & 16.407 \\
DarkIR + RL-AWB$^{\dagger}$ & 25.084 & 23.855 & \cellcolor{cthird}9.568 & 8.635 & 25.888 & 24.600 & \cellcolor{cthird}8.939 & 8.168 & 25.906 & 24.635 & \cellcolor{cthird}8.754 & 8.154 \\
\addlinespace[2pt]
MIRNet + DWB & 27.588 & \cellcolor{cthird}25.126 & 10.972 & \cellcolor{cthird}7.828 & 27.463 & 25.065 & 10.723 & 7.793 & \cellcolor{cthird}27.443 & 25.069 & 10.506 & 7.742 \\
MIRNet + FC4 & 25.765 & 21.254 & 24.441 & 15.714 & 26.287 & 21.501 & 23.775 & 15.509 & 26.373 & 21.354 & 24.222 & 15.849 \\
MIRNet + FFCC$^{\dagger}$ & 24.672 & 20.630 & 26.027 & 16.432 & 25.662 & 20.956 & 25.528 & 16.223 & 25.948 & 20.815 & 25.914 & 16.572 \\
MIRNet + RL-AWB & \cellcolor{cthird}27.681 & 24.790 & 14.989 & 10.177 & \cellcolor{csecond}28.665 & \cellcolor{csecond}26.273 & 10.836 & 8.020 & \cellcolor{csecond}29.040 & \cellcolor{csecond}26.564 & 9.903 & 7.653 \\
\addlinespace[2pt]
PromptNorm + DWB$^{\dagger}$ & 22.481 & 21.280 & 11.213 & 10.094 & 22.405 & 21.224 & 11.006 & 10.079 & 22.371 & 21.190 & 10.772 & 10.060 \\
PromptNorm + FC4$^{\dagger}$ & 23.939 & 20.520 & 23.044 & 15.431 & 24.474 & 20.888 & 22.734 & 15.146 & 24.330 & 20.674 & 23.377 & 15.520 \\
PromptNorm + FFCC$^{\dagger}$ & 26.197 & 20.068 & 25.623 & 17.081 & 26.692 & 20.223 & 25.075 & 16.830 & 26.762 & 20.085 & 25.570 & 17.178 \\
PromptNorm + RL-AWB$^{\dagger}$ & 25.978 & 24.519 & \cellcolor{csecond}8.788 & 8.077 & 26.768 & 25.243 & \cellcolor{csecond}8.138 & \cellcolor{csecond}7.592 & 26.768 & 25.243 & \cellcolor{cbest}8.023 & \cellcolor{cthird}7.583 \\
\hdashline
\addlinespace[2pt]
Ours (CTI-Retinexformer) & \cellcolor{csecond}27.969 & \cellcolor{cbest}25.953 & \cellcolor{cbest}8.340 & \cellcolor{cbest}6.958 & \cellcolor{cthird}27.779 & \cellcolor{cthird}25.867 & \cellcolor{cbest}8.023 & \cellcolor{cbest}6.900 & 27.421 & \cellcolor{cthird}25.540 & \cellcolor{csecond}8.062 & \cellcolor{cbest}7.056 \\
\bottomrule
\addlinespace[10pt]
\cmidrule[\heavyrulewidth]{1-5}
 & \multicolumn{4}{c}{Level 10} \\
\cmidrule(lr){2-5}
DSLR & PSNR$_L$ $\uparrow$ & PSNR$_C$ $\uparrow$ & MAE$_{\angle}$ $\downarrow$ & $\Delta$E$_{00}$ $\downarrow$ \\
\cmidrule{1-5}
Unprocessed & 26.631 & 22.830 & 20.891 & 13.061 \\
\cdashline{1-5}
\addlinespace[2pt]
TI-Retinexformer + DWB & 26.125 & 24.160 & 10.413 & 8.387 \\
TI-Retinexformer + FC4 & 26.694 & 21.072 & 23.793 & 15.810 \\
TI-Retinexformer + FFCC$^{\dagger}$ & 26.396 & 20.865 & 25.080 & 16.297 \\
TI-Retinexformer + RL-AWB$^{\dagger}$ & 25.919 & 24.801 & 8.664 & 7.985 \\
\addlinespace[2pt]
Retinexformer + DWB$^{\dagger}$ & 22.975 & 20.188 & 15.935 & 12.383 \\
Retinexformer + FC4 & 26.506 & 21.658 & 23.808 & 15.507 \\
Retinexformer + FFCC$^{\dagger}$ & 26.347 & 21.196 & 25.327 & 16.112 \\
Retinexformer + RL-AWB & \cellcolor{cbest}30.048 & \cellcolor{cbest}27.509 & 9.600 & \cellcolor{csecond}7.181 \\
\addlinespace[2pt]
DarkIR + DWB & 27.014 & 24.960 & 10.328 & 7.670 \\
DarkIR + FC4$^{\dagger}$ & 22.587 & 20.415 & 24.112 & 15.793 \\
DarkIR + FFCC$^{\dagger}$ & 26.859 & 21.354 & 25.040 & 15.961 \\
DarkIR + RL-AWB$^{\dagger}$ & 26.462 & 25.110 & \cellcolor{cthird}8.389 & 7.801 \\
\addlinespace[2pt]
MIRNet + DWB & 27.131 & 24.977 & 10.358 & 7.718 \\
MIRNet + FC4 & 26.795 & 21.641 & 23.581 & 15.444 \\
MIRNet + FFCC$^{\dagger}$ & 26.512 & 21.136 & 25.082 & 16.098 \\
MIRNet + RL-AWB & \cellcolor{csecond}29.124 & \cellcolor{csecond}26.734 & 9.423 & 7.411 \\
\addlinespace[2pt]
PromptNorm + DWB$^{\dagger}$ & 22.190 & 21.105 & 10.660 & 10.062 \\
PromptNorm + FC4$^{\dagger}$ & 24.658 & 20.917 & 22.882 & 15.173 \\
PromptNorm + FFCC$^{\dagger}$ & 26.912 & 20.310 & 24.738 & 16.713 \\
PromptNorm + RL-AWB$^{\dagger}$ & 27.147 & 25.571 & \cellcolor{cbest}7.724 & \cellcolor{cthird}7.297 \\
\cdashline{1-5}
\addlinespace[2pt]
Ours (CTI-Retinexformer) & \cellcolor{cthird}27.326 & \cellcolor{cthird}25.600 & \cellcolor{csecond}7.906 & \cellcolor{cbest}6.943 \\
\cmidrule[\heavyrulewidth]{1-5}
\end{longtable}

\end{document}